\documentclass[journal, 10pt]{IEEEtran}

\usepackage{xcolor}
\usepackage{booktabs}

\usepackage{caption}
\usepackage{subcaption}
\usepackage{threeparttable}

\usepackage{multirow}
\usepackage{tikz}
\usepackage{textcomp}
\usepackage{dsfont}
\usepackage{mathrsfs}
\usepackage[T1]{fontenc}
\usepackage{cite}
\usepackage{amsmath,amssymb,amsfonts,mathtools}
\usepackage{algorithm}
\usepackage{algcompatible}
\usepackage{graphicx}
\usepackage{textcomp}
\usepackage{wrapfig}
\usepackage{acronym}

\usepackage{url}

\acrodef{prop}[\textit{MIMORPH}]{MIMO Radio Platform for Heterogeneous wireless systems}
\acrodef{aae}[AAE]{Adversarial Autoencoder}
\acrodef{adc}[ADC]{Analog-to-Digital Converter}
\acrodef{aoa}[AoA]{Angle of Arrival}
\acrodef{aod}[AoD]{Angle of Departure}
\acrodef{ap}[AP]{Access Point}
\acrodef{cdf}[CDF]{Cumulative Distribution Function}
\acrodef{cef}[CEF]{Channel Estimation Field}
\acrodef{cfo}[CFO]{Carrier Frequency Offset}
\acrodef{cir}[CIR]{Channel Impulse Response}
\acrodef{cnn}[CNN]{Convolutional Neural Network}
\acrodef{csi}[CSI]{Channel State Information}
\acrodef{cs}[CS]{Compressed Sensing}
\acrodef{cnn}[CNN]{Convolutional Neural Network}
\acrodef{dft}[DFT]{Discrete Fourier Transform}
\acrodef{dl}[DL]{Deep Learning}
\acrodef{dust}[DUST]{Deep Unfolding Sparse Transformer mode}
\acrodef{edmg}[EDMG]{Enhanced Directional Multi Gigabit}
\acrodef{ekf}[EKF]{Extended Kalman Filter}
\acrodef{elu}[ELU]{Exponential-Linear Unit}
\acrodef{fmcw}[FMCW]{Frequency-Modulated Continuous-Wave}
\acrodef{fov}[FOV]{Field-of-View}
\acrodef{ft}[FT]{Fourier Transform}
\acrodef{har}[HAR]{Human Activity Recognition}
\acrodef{if}[IF]{Intermediate Frequency}
\acrodef{ifs}[IFS]{Inter-Frame Spacing}
\acrodef{iht}[IHT]{Iterative Hard-Thresholding}
\acrodef{ista}[ISTA]{Iterative Shrinkage-Thresholding Algorithm}
\acrodef{isac}[ISAC]{Integrated Sensing And Communication}
\acrodef{jcs}[JCS]{Joint Communication and Sensing}
\acrodef{los}[LOS]{Line-of-Sight}
\acrodef{liht}[LIHT]{Learned Iterative Hard-Thresholding}
\acrodef{lista}[LISTA]{Learned Iterative Shrinkage-Thresholding Algorithm}
\acrodef{mae}[MAE]{Mean Absolute Error}
\acrodef{mse}[MSE]{Mean Squared Error}
\acrodef{md}[$\mu$D]{micro-Doppler}
\acrodef{mimo}[MIMO]{Multiple Input Multiple Output}
\acrodef{mmwave}[mmWave]{Millimeter-Wave}
\acrodef{MUSIC}[MUSIC]{MUlti SIgnal Classification}
\acrodef{nlos}[NLOS]{Non-Line-of-Sight}
\acrodef{nn}[NN]{Neural Network}
\acrodef{ofdm}[OFDM]{Orthogonal Frequency Division Multiplexing}
\acrodef{omp}[OMP]{Orthogonal Matching Pursuit}
\acrodef{phy}[PHY]{Physical Layer}
\acrodef{pov}[PoV]{Point-of-View}
\acrodef{rf}[RF]{Radio Frequency}
\acrodef{relu}[ReLU]{Rectified Linear Unit}
\acrodef{rcs}[RCS]{Radar Cross-Section}
\acrodef{rss}[RSS]{Received Signal Strength}
\acrodef{rnn}[RNN]{Recurrent Neural Network}
\acrodef{rom}[ROM]{Read Only Memories}
\acrodef{rmse}[RMSE]{Root MSE}
\acrodef{sc}[SC]{Single Carrier}
\acrodef{sdr}[SDR]{Software Defined Radio}
\acrodef{siso}[SISO]{Single Input Single Output}
\acrodef{ssim}[SSIM]{Structural Similarity Index Metric}
\acrodef{sista}[SISTA]{Sequential ISTA}
\acrodef{slista}[SLISTA]{Sequential LISTA}
\acrodef{snr}[SNR]{Signal-to-Noise Ratio}
\acrodef{stft}[STFT]{Short Time Fourier Transform}
\acrodef{tf}[TF]{Time-Frequency}
\acrodef{toa}[ToA]{Time of Arrival}
\acrodef{lstm}[LSTM]{Long Short Term Memory}

\acrodef{osgr}[OSGR]{Open-set Gait Recognition}
\acrodef{orced}[OR-CED]{Open-set Recognition based on Contrastive constraint and Ensemble-based out-of-distribution Detection}
\acrodef{vae}[VAE]{Variational Auto-Encoders}
\acrodef{gan}[GAN]{Generative Adversarial Networks}
\acrodef{tcpcn}[TCPCN]{Temporal Convolution Point-Cloud Network}
\acrodef{elu}[ELU]{Exponential Linear Unit}
\acrodef{pcaa}[PCAA]{Point Cloud Adversarial Autoencoder}
\acrodef{mlp}[MLP]{Multi Layer Perceptron}
\acrodef{lgm}[L-GM]{Large margin Gaussian Mixture}

\graphicspath{{./figures/}}
\newcommand{\datasetname}{mmGait10}
\newcommand{\eq}[1]{Eq.~\eqref{#1}}
\newcommand{\fig}[1]{Fig.~\ref{#1}}
\newcommand{\tab}[1]{Tab.~\ref{#1}}
\newcommand{\secref}[1]{Section~\ref{#1}}
\newcommand{\alg}[1]{Alg.~\ref{#1}}

\newcommand{\rev}{}
\newcommand{\revv}{}

\newcommand{\mytexttilde}{{\raise.17ex\hbox{$\scriptstyle\mathtt{\sim}$}}}

\def\BibTeX{{\rm B\kern-.05em{\sc i\kern-.025em b}\kern-.08em
    T\kern-.1667em\lower.7ex\hbox{E}\kern-.125emX}}
    
\begin{document}

\title{Open-Set Gait Recognition from \\ Sparse mmWave Radar Point Clouds}

\author{Riccardo~Mazzieri,~\IEEEmembership{Graduate Student Member, IEEE}, Jacopo~Pegoraro,~\IEEEmembership{Member, IEEE}, Michele~Rossi,~\IEEEmembership{Senior Member, IEEE}
\thanks{Riccardo Mazzieri and Jacopo Pegoraro are with the Department of Information Engineering, University of Padova, Padova 35131, Italy (email: riccardo.mazzieri@phd.unipd.it; jacopo.pegoraro@unipd.it).}
\thanks{Michele Rossi is with the Department of Information Engineering, University of Padova, Padova 35131, Italy, and with the Department of Mathematics ``Tullio Levi-Civita'', University of Padova, Padova 35121, Italy (email: michele.rossi@unipd.it).}
\thanks{This work was supported by the European Union under the Italian National Recovery and Resilience Plan (NRRP) Mission 4, Component 2, Investment 1.3, CUP C93C22005250001, partnership on “Telecommunications of the Future” (Program “RESTART”) under Grant PE00000001.}
}

\maketitle

\begin{abstract}
The adoption of \ac{mmwave} radar devices for human sensing, particularly gait recognition, has recently gathered significant attention due to their efficiency, resilience to environmental conditions, and privacy-preserving nature.
In this work, we tackle the challenging problem of \ac{osgr} from sparse \ac{mmwave} radar point clouds. Unlike most existing research, which assumes a closed-set scenario, our work considers the more realistic open-set case, where unknown subjects might be present at inference time, and should be correctly recognized by the system.
Point clouds are well-suited for edge computing applications with resource constraints, but are more significantly affected by noise and random fluctuations than other representations, like the more common micro-Doppler signature. This is the first work addressing open-set gait recognition with sparse point cloud data. To do so, we propose a novel neural network architecture that combines supervised classification with unsupervised reconstruction of the point clouds, creating a robust, rich, and highly regularized latent space of gait features. To detect unknown subjects at inference time, we introduce a probabilistic novelty detection algorithm that leverages the structured latent space and offers a tunable trade-off between inference speed and prediction accuracy. Along with this paper, we release \texttt{mmGait10}, an original human gait dataset featuring over five hours of measurements from ten subjects, under varied walking modalities. Extensive experimental results show that our solution attains $24\%$ average F1-Score improvement over state-of-the-art methods \revv{adapted for point clouds}, across multiple openness levels.
\end{abstract}


\maketitle

\section{Introduction}
\label{sec:intro}
Radar technology based on \acp{mmwave} is being increasingly investigated for human sensing applications. This is due to its flexible and non-invasive nature and to its high sensing resolution due to the use of a high carrier frequency and a large bandwidth~\cite{zhang2023survey}. By leveraging the information in the backscattered electromagnetic signal, mmWave radars can infer key quantities such as the subject's position in the three-dimensional space and their radial velocity, with high precision.
\begin{center}    
    \includegraphics[width=\linewidth]{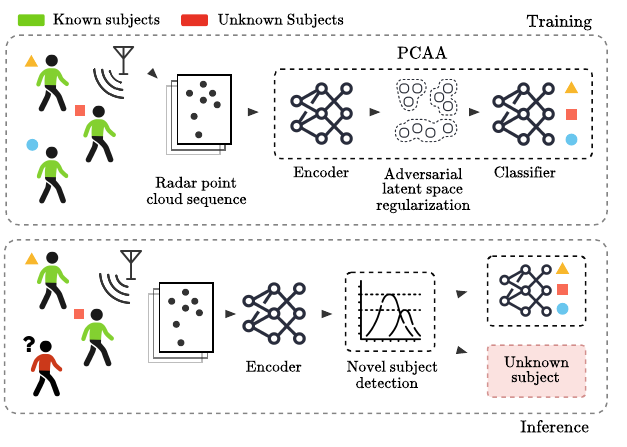}
\end{center}
Example applications include vital signs monitoring~\cite{li2009radar}, fall detection~\cite{li2022real}, human tracking~\cite{pegoraro2020multiperson} and real-time person identification~\cite{canil2021millitrace}. 

When considering person identification tasks, the adoption of radar technology entails a set of additional challenges with respect to more traditional approaches.
Indeed, differently from cameras, which leverage visual information to perform person identification tasks, radar sensing must solely rely on spatiotemporal features of the subjects' movements. These are often connected with the body shape and the way of walking, commonly referred to as \textit{gait}.

In the literature, gait is classified as a \textit{soft biometric}~\cite{nambiar2019gait}. Indeed, unlike \textit{hard biometrics} such as fingerprints or DNA, gait is subject to temporal variations and its discriminative power becomes weak when used with a large population. However, gait can serve as an effective discriminative feature in scenarios involving a limited number of subjects, at most in the range of a few tens~\cite{li2021mtpgait}. Example applications of remote gait recognition include medical diagnosis, home monitoring, user authentication, and surveillance systems~\cite{kong2024survey}.

The extraction of features of human gait is most commonly carried out by analyzing the so-called \textit{\ac{md} signature} of a subject. The micro-Doppler is a frequency modulation observed on the reflected signal, induced by the Doppler effect caused by small-scale movements of the target. As such, \ac{md} signatures contain time-frequency features of the subject's movements, which can be used to recognize an individual among a group. Typically, due to the complexity of the \ac{md} time-frequency patterns, \ac{md} signatures are processed using suitable (often \mbox{\ac{cnn}-based}) deep learning architectures that automatically extract gait features~\cite{pegoraro2020multiperson, seifert2019toward}. 

\ac{mmwave} radar-based person identification systems are especially appealing for distributed monitoring applications. In this domain, radar sensors are deployed at the edge of the communication network, equipped with low-power and resource-constrained computing devices~\cite{akan2020internet}. This poses two important research challenges.

First, collecting and processing radar data can be computationally demanding due to their large size and high resolution. This requires using radar data representations that trade some resolution or accuracy for an increased sparsity and lower size. A promising candidate in this sense is the point-cloud representation~\cite{camuffo2022recent}. By exploiting detection algorithms at the transmitter side, it is possible to infer the three-dimensional position and velocity of back-scattered points from the subject's surface and store them in sequences of collections of points, gaining access to spatiotemporal information regarding the subject movements. The point-cloud representation is widely used by LIDAR sensors, but it is difficult to process in the radar domain due to its sparsity, fast variability in time, and the small number of reflection points off the human body. 

Second, distributed monitoring applications are typically used in \textit{open-set} scenarios, where subjects that were not included in the training set may appear and disappear from the scene. In such setup, gait recognition methods must be able to effectively deal with \textit{unseen} subjects, by distinguishing them from those already known to the system. The problem of simultaneously classifying known subjects and detecting unknown ones using gait features is referred to in the literature as \ac{osgr}~\cite{yang2022multiscenario, moon2022open, geng2020recent}. Most existing works focus on the so-called \textit{closed-set} scenario, assuming that the same set of subjects present in the training set will remain unchanged during the test phase. However, for real applications, this assumption is quite restrictive, as the classifier would not be able to handle the presence of new subjects and would inevitably classify them as one of the already known ones. 

To the best of our knowledge, no previous work in the literature has jointly tackled both challenges, thus being unsuited for realistic edge-based people monitoring applications.

In this paper, we propose \ac{pcaa}, a novel deep learning architecture based upon the \ac{aae}~\cite{makhzani2015adversarial} that extracts meaningful gait features from sparse point-cloud data and uses them to perform classification of known subjects and detection of unknown ones during the testing phase. 
We remark that, differently from existing works that tackle \ac{osgr} from micro-Doppler gait signatures, our approach is the first one to specifically address the more challenging scenario where sparse radar point cloud sequences are used as input for the network. 

The main contributions of our work are:
\begin{enumerate}
    \item We propose and validate \ac{pcaa}, a novel neural network architecture capable of effectively classifying radar point clouds of human gaits. \ac{pcaa} minimizes a multi-task objective, effectively leveraging both label information and input data structure to build a highly regularized feature space.
    \item We design a simple and efficient algorithm to detect novel subjects from the extracted features at inference time, by successively addressing the \ac{osgr} problem. Additionally, our algorithm can also wait for a variable number of input steps before returning a prediction, thus obtaining a higher accuracy. This allows the user to tune inference speed for prediction accuracy.
    \item We acquire and publish \datasetname, a new high quality, multi-scenario, human-gait \ac{mmwave} radar dataset, consisting of approximately $5$ hours of measurements. 
    \revv{We release both \ac{md} spectrograms and pre-processed sparse point clouds} of ten subjects freely walking around within an indoor environment with different walking modalities.
    The large bandwidth and high angular resolution of the employed mmWave device allowed for denser point clouds as compared to those available from existing datasets, with point clouds containing an average of $150$ points per frame.
    \item We evaluate \ac{pcaa} in different experimental conditions and compare it with the latest available methods from the literature. We demonstrate that our method significantly outperforms existing solutions \revv{adapted for points clouds}, achieving $24\%$ average F$1$-Score improvement.
\end{enumerate}

The manuscript is structured as follows. In \secref{sec:related-work}, we provide a review of the existing literature on gait classification with mmWave radars, underlining the novel aspects of our work. In \secref{sec:method}, we thoroughly describe our method and the \ac{pcaa} architecture. In \secref{sec:dataset}, we provide details about the \datasetname{} dataset, from the experimental methodologies to statistics about the captured point clouds. In \secref{sec:results}, we show our experimental results. Concluding remarks are reported in \secref{sec:conclusion}.

\section{Related Work}
\label{sec:related-work}


\subsection{Deep Learning for point cloud gait recognition}
The problem of gait recognition from sparse radar point cloud sequences has been investigated by a few works in the recent literature~\cite{meng2020gait, pegoraro2021real, cheng2021person}. The most common approach involves the use of deep-learning-based feature extraction techniques.

In ~\cite{meng2020gait}, the authors introduce mmGaitNet, a deep neural network based on spatio-temporal convolutions. 
The multi-dimensional input point clouds are modeled as matrices, with each point's feature ($3$D Cartesian coordinates, velocity, and power) processed independently by a separate network branch. 
The obtained features are then processed by a convolutional network that performs feature fusion, followed by a fully connected neural network classifier.

In~\cite{pegoraro2021real}, the authors propose \ac{tcpcn}, a neural network model that processes the input data in two sequential steps. 
First, each point cloud in the input sequence is processed by a PointNet-based module~\cite{Qi_2017_CVPR}, obtaining a sequence of feature vectors of spatial and velocity information about the single point clouds. 
Then, a temporal convolution block based on causal dilated convolutions processes the sequence of feature vectors, therefore modeling temporal correlations.

The authors of~\cite{cheng2021person} tackle person re-identification, i.e., recognizing people across different environments from their gait. 
They propose SRPNet, a neural network that combines PointNet and a bidirectional Long Short Term Memory 
module to extract the spatio-temporal features from the input point cloud sequences. 

We stress that the above approaches are designed to solve the \textit{closed-set} gait classification problem. 
As such, they cannot solve the more challenging \ac{osgr} task, which requires identifying unknown subjects not present in the training set.

\subsection{Open-set gait recognition from radar measurements}
\label{sec:related-work-osgr}

Only a few studies have focused on the more general \ac{osgr} scenario. This task entails the following research challenges.
\begin{enumerate}
    \item \textit{Latent space regularization}: In closed-set classification, feature extraction is purely driven by supervised training to minimize the classification loss. 
    Therefore, the resulting latent space of feature representations strongly separates samples associated with different labels. 
    However, by only relying on the classification label information, the resulting distribution of feature representations will not be sufficiently representative of the \textit{general} spatio-temporal structure of the input, which may be critical to \ac{osgr}.
    To discriminate among seen and unseen subjects, some form of \rev{strong \textit{regularization} on the latent representations is needed, to achieve a structure which would not be possible to infer from the classification labels alone.}
    \item \textit{Novelty detection step}: to solve the \ac{osgr} task, the deep learning model must detect unseen subjects at inference time while classifying known subjects. 
    To solve this problem, a \textit{novelty detection} algorithm must be designed and applied to the observed samples at testing time. 
\end{enumerate}


\rev{Existing works tackle both challenges with a variety of approaches, ranging from generative models, contrastive learning strategies, or specialized loss functions.}

In~\cite{yang2019open}, the authors propose Open-\acs{gan}, a generative approach where a negative set of unknown classes is automatically generated during training by a \ac{gan} based generative model. 
In this way, the model learns a decision boundary between known and unknown samples.

\cite{ni2021open} presents a Deep Discriminative Representation Network (DDRN) trained with the cosine margin loss~\cite{wang2018cosface}. 
The latter promotes an increase in the inter-class distances and a reduction of the intra-class variance. 
The unknown detection is then performed by exploiting Extreme Value Theory (EVT), fitting the training feature vectors with a probability distribution to estimate the class inclusion probability. 
This approach allows finding a bound to the support region of the known classes in the embedding space, which is then used to recognize unknown subjects. 

\rev{In a later work, \cite{ni2022gait}, the same authors train an \ac{osgr} network through a \ac{lgm} loss, which shapes the resulting latent space of gait features as a Mixture of Gaussians distribution where each subject is modeled as a different Gaussian component. By exploiting this property of the feature space, known users are directly classified according to the class-posterior probability, while unknown ones are identified by setting a probability threshold.} 

In~\cite{yang2022multiscenario}, \ac{orced}, a dual branch neural network based on the ladder \ac{vae} architecture is proposed. 
\ac{orced} performs both reconstruction and classification of radar \ac{md} signatures. 
To counteract the negative effect of samples that are hard to discriminate, the authors introduce a contrastive loss term to force latent representations of each class to be close in the latent space. 
During inference, the distributions of the reconstruction errors and latent representations of training samples are modeled as univariate and multivariate Gaussians, respectively. 
Then, two out-of-distribution detectors are jointly used to separate known and unknown subjects. 

All the above-mentioned works use radar \ac{md} spectrograms to capture the type of action or gait of different subjects. 
However, to the best of our knowledge, no works have focused on deep learning systems capable of performing \ac{osgr} directly on radar point cloud sequences. 

Extracting features from dynamic, sparse radar point clouds requires the adoption of specialized neural architectures capable of extracting features from individual multidimensional point clouds and their evolution in time. 
While neural architectures for \ac{md} signatures are largely taken from the vast image processing literature, advanced neural networks for radar point cloud sequences are less investigated.

Furthermore, the lack of well-established decoder architectures for point cloud data makes unsupervised feature learning extremely challenging. Several ad-hoc decoder architectures have been proposed for point cloud completion and reconstruction~\cite{yang2018foldingnet, huang2020pf}. 
However, these mostly exploit the local structure typical of very dense and clean point cloud data, such as the ones produced by LIDAR sensors, making their use impractical for noisy and sparse radar point clouds. 

\subsection{Background on radar point clouds}
\label{sec:radar_background}

A \ac{mimo} \ac{fmcw} radar 
works by transmitting bursts of \textit{chirp} signals, linearly sweeping a bandwidth $B$. Chirp bursts, which constitute a \textit{radar frame}, are repeated with period $T_f$ seconds. The reflected copies of the signal from the environment are collected and processed by a receiver antenna array. Radar signal processing algorithms allow the joint estimation of the distance, the radial velocity, and the angular position of the targets~\cite{patole2017automotive}.
Distance and velocity are obtained by computing the frequency shift induced by the delay of each reflection, through conventional \textit{range-Doppler} processing~\cite{richards2010principles}.
To obtain a sparse set of dominant reflection points from the raw radar signal, range-Doppler processing is usually followed by detection algorithms such as constant false-alarm rate~\cite{richards2010principles}.
Then, the use of multiple receiving antennas allows obtaining the \ac{aoa} of the reflections along the azimuth and the elevation dimensions. This is done by leveraging the phase shifts measured by the different antenna elements of a planar antenna array. 
This returns a set of points, termed radar point cloud, that can be transformed into the $3$-dimensional Cartesian space ($x-y-z$) using the distance, azimuth, and elevation angles information of the reflectors. The velocity information is also appended to the components of each point, making it 4-dimensional.

\section{Proposed Method}
\label{sec:method}

In this work, we tackle the problem of \ac{osgr} from sparse radar point clouds. 
Our system consists of two main phases: \textit{deep feature extraction} and \textit{unknown subject detection}.
In the first phase, the system extracts compact features of human gait from the input radar point cloud sequences. To this end, we propose \ac{pcaa}, a deep neural network that reaps the benefits of supervised and unsupervised learning strategies. In \secref{subsec:preprocessing} we describe the pre-processing steps that are applied to the raw radar point clouds before they are fed to the network. Then, the network architecture and training strategy are thoroughly described in \secref{subsec:model-architecture} and \secref{subsec:model-training} respectively.

In the second phase, we leverage the regularized structure of the trained latent space to perform the novel subject detection task. To this end, we propose a simple detection algorithm, capable of determining whether an input gait feature vector belongs to a known or unknown subject, at inference time. The proposed novel subject detection algorithm is described in \secref{subsec:ood-detection}.

\subsection{Point cloud pre-processing}
\label{subsec:preprocessing}
Before being fed to our model, data collected by the radar undergoes a series of preprocessing steps. 
The first step takes the raw radar data and applies detection, clustering, and tracking algorithms to isolate the points reflected off the subject's body from the ones due to spurious reflections from the environment. 
To this end, we employ the same procedure proposed in~\cite{canil2021millitrace}, which applies a Kalman Filter to track the point cloud as it moves in space, coupled with a clustering algorithm to select the points corresponding to the subjects. 
In the following, we define a single reflected point as a vector \mbox{$\mathbf{x} = \left[x, y, z, v\right]^{\mathsf{T}}$}, including its spatial coordinates $x,y,z$, and its radial velocity $v$. We formally define a point cloud as a \mbox{set $\mathcal{X}$}, containing a variable number of points.

To build the input for our neural network, we split each point cloud sequence measurement in windows of three seconds, corresponding to $N_f$ frames taken with step size $s$. 
Additionally, since the our architecture requires a fixed number of points for each point cloud, we apply a pre-processing strategy to enforce $|\mathcal{X}| = N_p$. 
In the case where $|\mathcal{X}| > N_p$, we randomly remove the excess points, while if $|\mathcal{X}| < N_p$, we randomly repeat points until we reach the desired amount. 
Our choice for the value of $N_p$ was taken by observing the average number of captured points from the radar device and the clustering procedure. 
We then subtract the mean of each feature dimension from each point, so that each point cloud is centered in the origin of the coordinate system. 
We denote the resulting sequence of normalized point clouds, with a fixed number of points, by $\mathbf{X}_{1:N_f} = \{\mathbf{X}_t\}_{t=1}^{N_f}$, where $\mathbf{X}_t \in \mathbb{R}^{N_p \times 4}$. 

We denote by $\mathbf{e}_i$ the one-hot encoded vector of all zeros but the $i$-th component which equals $1$, and we define \mbox{$S=\{1, 2, \dots, M\}$} as the set of subject IDs.
By applying the above procedure to all our measurements we build a dataset of input-label pairs of the form $(\mathbf{X}_{1:N_f}, \mathbf{y})$, where \mbox{$\mathbf{y} \in \{\mathbf{e}_i\}_{i\in S}$}.

To tackle the \ac{osgr} task, we consider a dataset divided into a set of known subjects, to use for training, and a set of unknown subjects, never seen by the model during training. 
These are used to assess the ability of the network to recognize novel subjects. 
We define \mbox{$S_K \subset S$} the set of known subject IDs and \mbox{$S_U \subset S$} the set of the unknown ones, so that $\{S_K, S_U\}$ is a partition of $S$. The cardinalities of $S_K$ and $S_U$ depend on the specific scenario. Following the \ac{osgr} literature, we measure how many unknown subjects are present with respect to the known ones using the \textit{openness} metric, whose definition and impact on the performance of \ac{pcaa} are presented in \secref{sec:evaluation-metrics}. 

Using the above partition, we split the whole dataset into the known partition $\mathcal{K}$ and unknown partition $\mathcal{U}$,  which are defined as follows
\begin{align}
    \mathcal{K} = & \{(\mathbf{X}_{1:N_f},\mathbf{y}): \ \mathbf{y} \in \{\mathbf{e}_i\}_{i \in S_K}\},\\
    \mathcal{U} = & \{(\mathbf{X}_{1:N_f},\mathbf{y}): \ \mathbf{y} \in \{\mathbf{e}_i\}_{i \in S_U}\}.
\end{align}
The above sets of point cloud sequences are then used to train the proposed \ac{pcaa} model as detailed in the next section.

\subsection{Model architecture}
\label{subsec:model-architecture}

\ac{pcaa} leverages the strength of both supervised and unsupervised learning to extract meaningful features from radar point clouds, which is the first necessary step to solve the \ac{osgr} task. Our architectural design is based upon the \ac{aae}, a hybrid neural network architecture inspired by both \ac{vae}s and \ac{gan}s~\cite{makhzani2015adversarial}. 

The \ac{aae} architecture has two main components, an autoencoder module and a discriminator module. 
Let $\mathbf{X}_{1:N_f}$ be an input point cloud sequence, $\mathbf{z}$ its latent representation computed by the encoder module, and $\mathbf{\hat{X}}_{1:N_f}$ the reconstruction obtained from the decoder. We define as $p_d(\mathbf{X}_{1:N_f})$ the data distribution, by $q(\mathbf{z}|\mathbf{X}_{1:N_f})$ the encoding distribution induced by the encoder, and by $p(\mathbf{\hat{X}}_{1:N_f}|\mathbf{z})$ the distribution of the reconstructed point cloud sequence produced by the decoder. The core idea of the \ac{aae} is to match the aggregated encoder output distribution over all possible input point cloud sequences, $q(\mathbf{z})$, to a prior distribution $p(\mathbf{z})$ that can be specified during the design of the system.
To this end, a \textit{Discriminator} module is trained to predict whether its input was generated by the encoder module or sampled from the prior distribution. Jointly optimizing the encoder and the discriminator will induce the encoder to accurately reconstruct the input and to produce an aggregated posterior encoding distribution $q(\mathbf{z})$ that matches the prior $p(\mathbf{z})$.
It has been shown that the \ac{aae} architecture obtains more general and robust feature representations of the input when compared to classic approaches based on the \ac{vae}.
Some experimental results have also been shown for point cloud data~\cite{makhzani2015adversarial, zamorski2020adversarial}. 
Differently from \ac{vae}, no restrictions are imposed on the type of prior distribution $p(\mathbf{z})$ since the model only \textit{samples} from it.

\begin{figure*}
    \centering
    \includegraphics[width=\linewidth]{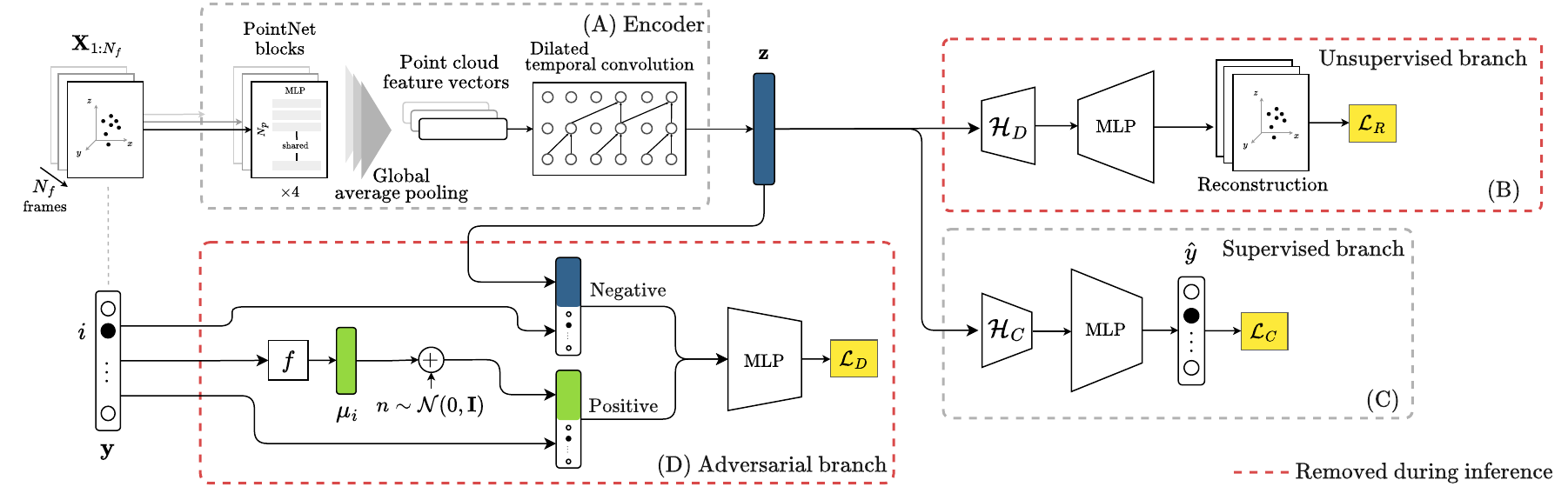}
    \caption{The figure depicts the proposed architecture and the flow of computations, going from left to right. The input point cloud sequence $\mathbf{X}_{1:N_f}$ is first fed through the \mbox{Encoder module (A)}, which outputs a latent representation $\mathbf{z}$. This is fed to the \mbox{Unsupervised branch (B)}, which solves a point cloud sequence reconstruction task, and to the \mbox{Supervised branch (C)}, which solves the closed-set subject ID classification task. The \mbox{Adversarial branch (D)} enforces the desired aggregated posterior distribution $q(\mathbf{z})$ of the latent representations. In our case, this corresponds to a multimodal Gaussian distribution. The modules outlined in red are employed only during training and removed during inference. }
    \label{fig:model-architecture}
\end{figure*}

We now describe each part of our model in detail, referring to the diagram of the \ac{pcaa} architecture reported in \fig{fig:model-architecture}.

\noindent \textbf{(A) Encoder:}
The Encoder block of our network processes the input point clouds to extract compact spatio-temporal features. To this end, we adapt the \ac{tcpcn} architecture proposed in~\cite{pegoraro2021real}, which was originally designed for gait classification from radar point clouds (see \mbox{\secref{sec:related-work-osgr}}). 

As a first step, each point cloud of the input sequence is processed by $4$ consecutive network blocks inspired by the PointNet architecture~\cite{Qi_2017_CVPR}, which we refer to as PointNet blocks. 
In the $i$-th block, the points of each point cloud in the sequence are processed in parallel by the \textit{same} MLP with $n_i$ output neurons. This is then followed by batch normalization~\cite{ioffe2015batch} and an \ac{elu}~\cite{clevert2015fast} activation function, resulting in a new set of $N_p$, $n_i$-dimensional points as the output. 
In our implementation, we choose \mbox{$n_1=n_2=n_3=512$} and \mbox{$n_4=1024$}. 
After the last PointNet block, a global average pooling operation is applied at each point cloud of the sequence, to aggregate the $N_p$ feature vectors into a single one of dimension $1024$. 
This step is essential to make the network structure invariant to random permutations of input points, which is a key property of
neural networks for point cloud processing. 
The resulting output is a sequence of feature vectors, one per point cloud in the sequence, which contain information regarding the position of points and their velocity.

Next, a temporal convolution block is employed to process the temporal correlations in the point cloud sequences. Specifically, $6$ layers of time-dilated, one-dimensional causal convolutions are employed~\cite{yu2015multi} along the time dimension. Different from standard convolutions, dilated convolutions adopt a filter with gaps (dilations) between filter elements. 
This makes it possible to expand the receptive field without increasing the number of parameters, enabling the network to capture multi-scale contextual information more efficiently.
The convolutional layers use kernels of length $3$, with dilations $(1,2,4,1,2,4)$, and filter sizes $(16,32,64,128,512)$. After a final average pooling operation along the time dimension and a single linear projection layer, the Encoder module returns the latent representation vector of the whole point cloud sequence, denoted by $\mathbf{z} \in \mathbb{R}^{K}, K=32$.

\noindent \textbf{(B) Unsupervised branch:}
This part of the network is responsible for the reconstruction of the input point cloud sequence from the compressed representation vector $\mathbf{z}$. 
Note that we are not interested in the reconstructed point cloud sequence per-se. 
However, this branch is essential to extract meaningful features that contain general information about the subject's movement pattern, beyond that related to the classification label.

In the Unsupervised branch, we introduce a \textit{projection head}~\cite{chen2020simple}, $\mathcal{H}_D$, to learn a transformation of $\mathbf{z}$ that specializes in reconstructing the input. $\mathcal{H}_D$ is a fully connected layer with $64$ neurons, equipped with an \ac{elu}. 
Then, the Decoder block 
performs a series of upsampling steps to yield an output tensor of the same shape as the input. 
Its architecture consists of a simple dense MLP with $5$ hidden layers with $(1125, 2250, 4500, 9000, 18000)$ neurons, each followed by an \ac{elu} non-linearity. 
The last layer reshapes the $18000$ outputs into the reconstructed point cloud sequence of shape $(4, 30, 150)$.
Note that, unlike most autoencoder architectures, we do not adopt a symmetrical structure with respect to the encoder module (i.e. convolutions followed by deconvolutions for \ac{cnn}-based autoencoders). 
The choice of this simple MLP structure is due to the non-invertibility of the PointNet module due to the global average pooling operation, as discussed in \secref{sec:related-work-osgr}.

\noindent \textbf{(C) Supervised branch:}
This branch is designed to perform subject classification from the compressed representation $\mathbf{z}$. Similarly to $\mathcal{H}_D$, we use a projection head, $\mathcal{H}_C$, to transform $\mathbf{z}$ into a new representation that specializes in classification. 
We found using projection heads to be beneficial to learn a more general representation $\mathbf{z}$, as motivated in~\cite{chen2020simple}.
After the projection head, a single output layer with a softmax activation function performs the classification task by returning a probability distribution over the classes.

\noindent \textbf{(D) Adversarial branch:}
The \ac{aae} architecture leverages a discriminator module to encourage the encoder to produce latent representation distributed according to some prior distribution. 

Given the input tuple \mbox{$(\mathbf{X}_{1:N_f},\mathbf{e}_i)$}, belonging to the known dataset partition $\mathcal{K}$, the key idea is to constrain the encoder to generate a representation $\mathbf{z}$ according to a Gaussian distribution $\mathcal{N}(\boldsymbol{\mu}_i,\mathbf{I}_K)$, where $\mathbf{I}_K$ is the $K \times K$ identity matrix. 
In simpler terms, our primary aim is to shape the probability distribution induced by the encoder, with conditioning arising from label information, leading to samples of different classes being distributed in separate clusters in the latent space.
This choice grants both good separability of the classes, to reach high classification accuracy in the closed set, and \textit{soft transitions} between clusters, learning smooth semantic representations.
\rev{Notably, the adversarial and unsupervised branches are only employed during training and are discarded at inference time, ensuring a low memory footprint suitable for edge devices.}

Differently from other approaches that leverage conditional information from labels~\cite{sun2020conditional}, we choose to fix the individual means, $\boldsymbol{\mu}_i$, of the Gaussian distributions, instead of learning them along with the network's weights. Specifically, we define a one-to-one mapping $f$ between each possible one-hot encoded label, $\mathbf{e}_i$, and the corresponding vector, $\boldsymbol{\mu}_i \in \mathbb{R}^K$, such that $\| \boldsymbol{\mu}_i - \boldsymbol{\mu}_j\|_2 > M$ and $\boldsymbol{\mu}_i \in \mathcal{S}_{M}$, where $\|\cdot\|_2$ is the euclidean norm and $\mathcal{S}_{M}$ is the $K$-dimensional hypersphere of radius $M$. This approach allows carefully designing \textit{where} the distributions are placed in the latent space, ensuring that all the space is effectively exploited.
Hence, it allows the network to only focus on \textit{how} to distribute the latent representations in a smaller region of the latent space.

After computing $\boldsymbol{\mu}_i$, we sample from distribution $\mathcal{N}(\boldsymbol{\mu}_i,\mathbf{I}_K)$, concatenate the result with the one-hot encoded label vector and input it to the discriminator as a \textit{positive} example, that is, a sample coming from the prior distribution $p(\mathbf{z})$. At the same time, the concatenation of the latent $\mathbf{z}$ with the one-hot label vector is provided as a \textit{negative} example, i.e., a sample coming from the aggregate posterior probability generated by the encoder $q(\mathbf{z})$. A similar concatenation procedure is used in the semi-supervised \ac{aae} variation presented in ~\cite{makhzani2015adversarial}, where the one-hot label acts as a switch allowing the Discriminator to make its decision according to the class information. The Adversarial module finally outputs the probability that the input is generated from the prior distribution.

\subsection{Model training}
\label{subsec:model-training}

The training procedure of \ac{pcaa} consists of two sequential optimization stages, namely \textit{feature extraction} and \textit{regularization}. 
In the feature extraction step, the Encoder, the Supervised, and the Unsupervised branches are jointly optimized. For the supervised component, the cross-entropy loss function between the predicted class probabilities  $\hat{\mathbf{y}}$ and the one-hot encoded label $\mathbf{y}$ is employed
\begin{equation}
\label{eq:ce-loss}
\mathcal{L}_C = - \sum_{i=1}^{M} \mathbf{y}_i \log \left(\hat{\mathbf{y}}_i\right).
\end{equation}

The training criterion for the unsupervised component is instead defined as the average Chamfer distance between the reconstructed and original point cloud sequences. Given an input point cloud sequence $\mathbf{X}_{1:N_f}$, we will denote the $i$-th point cloud of the sequence as $\mathbf{X}_{i}$. The Chamfer distance between $\mathbf{X}_{1:N_f}$ and its reconstruction $\hat{\mathbf{X}}_{1:N_f}$ is defined as
\begin{equation}
\label{eq:chamfer-loss}
\mathcal{L}_R =\frac{1}{N_f} \sum_{i=1}^{N_f} \left( \sum_{\mathbf{x} \in \mathbf{X}_{i}} \underset{\hat{\mathbf{x}} \in \hat{\mathbf{X}}_i}{\min} \|\mathbf{x}-\hat{\mathbf{x}}\|_2^2
+
\sum_{\hat{\mathbf{x}} \in \hat{\mathbf{X}}_i} \underset{\mathbf{x} \in \mathbf{X}_i}{\min} \|\mathbf{x}-\hat{\mathbf{x}}\|_2^2
\right).
\end{equation}
Given two point clouds, \eq{eq:chamfer-loss} sums \textit{(i)} the squared distance between each element in the first point cloud and its nearest neighbor in the second point cloud, and \textit{(ii)} the squared distance between each element in the second point cloud and its nearest neighbor in the first point cloud. 
In \eq{eq:chamfer-loss}, the original point cloud $\mathbf{X}_{i}$ assumes the role of the first point cloud, while $\hat{\mathbf{X}}_i$ is the second one.
The Chamfer distance is widely used for point cloud reconstruction tasks, due to its differentiability and computational efficiency \cite{camuffo2022recent}, \cite{wu2021densityaware}.
The final optimization criterion for the first stage is defined as
\begin{equation}
\label{eq:joint-loss}
\mathcal{L} = \mathcal{L}_R + \mathcal{L}_C.
\end{equation}

By minimizing \eq{eq:joint-loss}, \ac{pcaa} learns to extract features that account for both class information and intrinsic patterns in the data, such as the walking style or body features of the subjects. This is essential to obtain general semantic features, to be used later for the detection of different new subjects.

In the regularization stage, the parameters of the Discriminator are updated according to a Wasserstein criterion with gradient penalty~\cite{gulrajani2017improved}
\begin{align}
\label{eq:wasserstein-loss}
     \mathcal{L}_D = &\mathbb{E}_{q(\mathbf{z})}[D(\mathbf{z})] - \mathbb{E}_{p(\mathbf{z^*})}[D(\mathbf{z^*})] \\
     &+ \lambda \mathbb{E}_{u(\hat{\mathbf{z}})}\left[(\| \nabla_{\hat{\mathbf{z}}} D(\hat{\mathbf{z}}) \|_2 -1)^2\right],
\end{align}
where $\mathbb{E}_{f(\cdot)}$ is the expectation under the probability density function $f$, $u(\hat{\mathbf{z}})$ is a uniform distribution along lines connecting couples of points sampled from $q(\mathbf{z})$ and $p(\mathbf{z})$.
We adopt the Wasserstein criterion coupled with gradient penalty instead of the standard \ac{gan} loss, due to its improved convergence properties and training stability.

We remark that the training phase is only carried out using the known dataset $\mathcal{K}$. More in detail, as it is a standard practice in deep learning, we divided $\mathcal{K}$ into random training, validation, and test splits with a $(0.8, 0.1, 0.1)$ ratio.

For both stages of training, we employed the Adam optimizer~\cite{kingma2014adam} with learning rate $\eta=10^{-4}$. All the models are implemented in PyTorch~\cite{paszke2019pytorch} and trained on an NVIDIA RTX 3080 GPU. The code implementation is publicly available at \url{https://github.com/rmazzier/OpenSetGaitRecognition_PCAA}. We summarize all the relevant model parameters in \tab{tab:model-params}.

\begin{table}[t!]
\small
\centering
\caption{Summary of the system hyperparameters}
\label{tab:model-params}
\begin{tabular}{@{}lcr@{}}
\toprule
\multicolumn{3}{c}{\textbf{System hyperparameters}} \\ \midrule
Total number of subjects & $M$ & 10 \\
Max. number of points & $N_p$ & 150 \\
Input point cloud frames & $N_f$ & 30 \\
Crop step size & $s$ & 6 \\
Latent space dimension & $K$ & 32 \\
Gradient penalty weight & $\lambda$ & 15 \\
\bottomrule
\end{tabular}
\end{table}

\subsection{Unkown subject detection}
\label{subsec:ood-detection}

To distinguish known subjects from unknown ones during inference, we propose a novel subject detection algorithm based on the statistical properties of the feature space learned by \ac{pcaa}. The algorithm relies on the strong regularization capabilities of the \ac{aae} architecture.

Specifically, we exploit the fact that, after training, feature vectors extracted from radar traces of subject $i$ follow a normal distribution $\mathcal{N}(\boldsymbol{\mu}_i,\mathbf{I}_K)$, where the mean vectors $\{\boldsymbol{\mu}_i\}_{i=1}^{M}$ were deterministically chosen before training according to the procedure described in \secref{subsec:model-architecture}. 
We can therefore define the \textit{known distribution}, as a mixture of Normal Gaussian distributions with equal weights, which is written as
\begin{equation}
\label{eq:likelihood-score}
    p(\mathbf{z}) = \frac{1}{M} \sum_{i=1}^{M} \frac{\exp\left(-\frac{1}{2} (\mathbf{z} - \boldsymbol{\mu}_i)^{\mathsf{T}} \mathbf{I}^{-1} (\mathbf{z} - \boldsymbol{\mu}_i) \right)}{\sqrt{(2\pi)^K|\mathbf{I}_K|}},  
\end{equation}
where $\mathbf{I}$ is the identity covariance matrix and $\cdot^{\mathsf{T}}$ denotes matrix and vector transposition.
Our choice of assigning equal weights of $1/M$ to each mode of the Gaussian distribution represents the prior belief that each subject will be equally likely to appear in the test data distribution. 
In scenarios where this assumption does not hold, a different configuration of mode weights can be adopted.
Given a feature vector $\mathbf{z}$ at inference time, we can use $p(\mathbf{z})$ as a score that quantifies the likelihood of that vector belonging to the known distribution.
Our novel subject detection strategy exploits the sequential nature of the input data.
Rather than immediately classifying an observed input sample $\mathbf{X}_{1:N_f}$, we base the final decision on the observation of a sequence $\{\mathbf{X}_{1:N_f}[t]\}_{t=1}^{k}$ of input point cloud data, where the number of samples in the sequence, $k$, is chosen as a system hyper-parameter.

We report the pseudocode of the proposed novel subject detection algorithm in \alg{alg:novel-subject-algorithm}. In lines $1$ through $5$, we loop through the sequences $\mathbf{X}_{1:N_f}[t]$, for $t=1,\dots,k$, and for each one we compute the closed-set prediction $\hat{y}[t]$ and a score $s[t]=p(\mathbf{z}[t])$, thus obtaining two sequences $\{\hat{y}[t]\}_{t=1}^k$ and $\{s[t]\}_{t=1}^k$. Then in lines 6 through 10 we check if more than half of the scores in $\{s[t]\}_{t=1}^k$ are greater than $\tau_p$. If so, we exit the procedure, returning the most frequent prediction, otherwise, the subject is classified as unknown. 
The threshold $\tau_p$ is computed by applying a secondary procedure. First, we sample a random subject identifier $c \in S_{U}$ and define \mbox{$\mathcal{U}^*=\{(\mathbf{X}_{1:N_f}, \mathbf{y}) : \mathbf{y} = \mathbf{e}_c\} \subset \mathcal{U}$}. We then compute the scores, as defined in \alg{alg:novel-subject-algorithm}, of all samples in $\mathcal{K}$ and $\mathcal{U}^*$. Finally, we set $\tau_p$ to the threshold between the two score distributions that maximizes the true positive rate and minimizes the false positive rate.  

In practice, the choice of the threshold $\tau_p$ requires the presence of only one additional subject outside of $S_K$, which can be seen as an additional requirement for system calibration. In our simulations, we draw it from the set of unknown subjects, and then run the final novel subject detection procedure on the newly defined set of unknown subjects $\mathcal{U} \setminus \mathcal{U^*}$.

\begin{algorithm}[t!]
\small
	\caption{Novel subject detection procedure}
	\label{alg:novel-subject-algorithm}
	\begin{algorithmic}[1]
		\REQUIRE Sequence of point cloud sequences \mbox{$\{\mathbf{X}_{1:N_f}[t]\}_{t=1}^{k}$}, threshold $\tau_p$, trained encoder $\mathcal{E}$, projection head $\mathcal{H}_C$ and classifier $\mathcal{C}$.
		\ENSURE Final \ac{osgr} prediction $\hat{y}_{OS}$.

\FOR{$t=1,2,\dots,k$}
\STATEx \textcolor{gray}{\texttt{// Compute feature vector}}
\STATE $\mathbf{z}[t] \leftarrow \mathcal{E}(\mathbf{X}_{1:N_f}[t])$
\STATEx \textcolor{gray}{\texttt{// Compute closed-set prediction}}
\STATE $\hat{y}[t] \leftarrow \mathcal{C}(\mathcal{H}_{C}(\mathbf{z}[t]))$
\STATEx \textcolor{gray}{\texttt{// Compute score}}
\STATE $s[t] \leftarrow p(\mathbf{z}[t])$ \eqref{eq:likelihood-score}
\ENDFOR
\STATEx \textcolor{gray}{\texttt{// Check how many scores are above the threshold}}
\IF{$| \{s[t] > \tau_p, t=1,\dots,k \} |>k/2$} 
\STATE $\hat{y}_{OS} \leftarrow$ Most frequent prediction in $\{\hat{y}[t]\}_{t=1}^k$
\ELSE 
\STATE $\hat{y}_{OS} \leftarrow$ Unknown subject
\ENDIF
	\end{algorithmic}
\end{algorithm}

\section{mmGait10 dataset Structure}
\label{sec:dataset}

To evaluate our approach, we collect mmGait10, a radar dataset for human gait analysis \revv{\cite{mazzieri_2025_14974386}, that includes both \ac{md} spectrograms and point cloud sequences}.
\revv{This comprises approximately 5 hours of radar measurements collected from $10$ subjects}. The participants have a diverse range of physical characteristics, including sex, height, and weight, to capture a realistic variety of walking patterns. Informed consent was obtained from all the involved subjects before data collection.
More details regarding the amount of recorded data for each subject are reported in \tab{tab:subj-dataset-stats}. 
Each subject is instructed to walk freely along random trajectories in a $7.81$~m $\times$ $7.26$~m room. 
To further enhance diversity and realism, each subject's gait is recorded under three different conditions: \textit{i)} walking freely, \textit{ii)} walking freely while holding a smartphone, and \textit{iii)} walking freely with their hands in their pockets. 
These different walking modalities, where the motion of different body parts is present or absent (e.g., when holding a smartphone, the arm's natural swinging motion is restricted), enable learning more general and robust gait features.
Each walking modality is carried out for approximately $10$ minutes, for a total recording time of approximately $30$ minutes per subject.
All the measurements are collected with a commercial Texas Instruments MMWCAS-RF-EVM FMCW Radar, operating in the \mbox{$77$-$81$~GHz} frequency band at a frame rate of $10$~Hz. 
We report the radar parameters and the resolution of our measurements in \tab{tab:radar-params}.

\begin{table}[]
\small
\centering
\caption{Summary of the radar parameters}
\label{tab:radar-params}
\begin{tabular}{@{}lr@{}}
\toprule
\multicolumn{2}{c}{\textbf{Radar Parameters}} \\ \midrule
No. of TX Antennas & 12 \\
No. of RX Antennas & 16 \\
No. of Virtual Antennas & 192 \\
Start frequency & 76 GHz \\
Chirp bandwidth & 2.56 GHz \\
Max velocity & 2.68 m/s \\
Max range & 15.00 m \\
Range resolution & 5.86 cm \\
Velocity resolution & 4.2 cm/s \\
Angle resolution & 0.69 deg \\ \bottomrule
\end{tabular}
\end{table}


The point clouds obtained from the measurements are filtered to remove clutter due to reflections from the environment, hence only retaining the points reflected from the subjects' bodies. 
To this end, we apply the procedure used in~\cite{canil2021millitrace}, where a Kalman filter-based algorithm is employed to track the subject in the room, and a density-based clustering procedure performs the clutter removal, selecting only the points of interest.

\begin{figure}
    \centering
    \includegraphics[width=\linewidth]{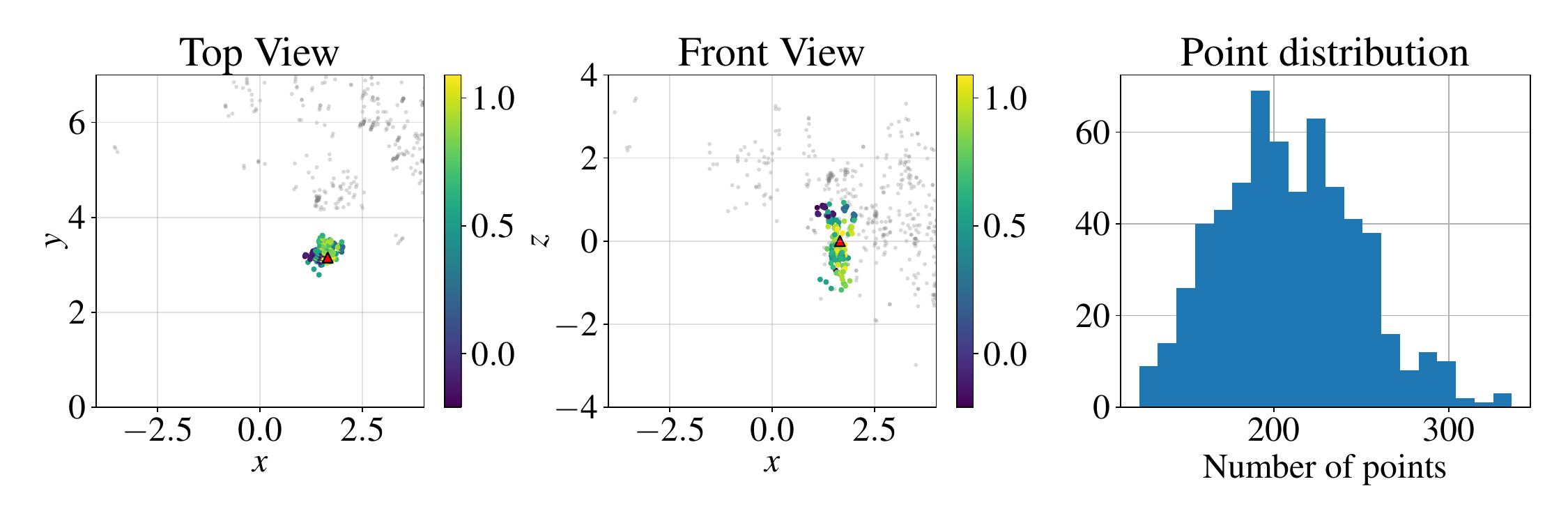}
    \caption{Example radar point cloud from the proposed dataset. In the two leftmost plots, the color coding of each point denotes their radial velocity, while the gray dots denote background reflections. On the right, we show an example distribution of the number of points in the point clouds.}
    \label{fig:dataset-example}
\end{figure}

The density of the recorded point clouds is a critical factor for the success of \ac{osgr}. The choice of the device together with an accurate tuning of the sensing parameters allowed us to obtain denser point clouds with respect to those of existing publicly available datasets, with an average of $200$ points for each frame. \fig{fig:dataset-example} shows an example of an extracted point cloud reflected off the human subject, together with the distribution of the number of points from an example sequence.

\begin{table} [t!]
\small
	\begin{center}
		\begin{tabular}{ccccc}
\toprule
& & \multicolumn{3}{c}{\textbf{No. of frames}} \\
\cmidrule(lr){3-5}
\textbf{Subject} & \textbf{Gender} & \textbf{Mod. 1} &\textbf{Mod. 2} & \textbf{Mod. 3} \\
\midrule
0 & M & $6061$ & $5587$ & $5020$ \\
1 & M & $5995$ & $6011$ & $5819$ \\
2 & M & $6463$ & $5624$ & $5937$ \\
3 & F & $5993$ & $6181$ & $6234$ \\
4 & M & $5375$ & $5763$ & $5829$ \\
5 & M & $6625$ & $6291$ & $6217$ \\
6 & M & $6289$ & $5984$ & $6444$ \\
7 & F & $6069$ & $6000$ & $6339$ \\
8 & M & $6305$ & $6398$ & $6466$ \\
9 & M & $6957$ & $6036$ & $6258$ \\

\bottomrule
\end{tabular}
	\end{center}
	\caption{Number of frames per subject in the dataset.} 
	\label{tab:subj-dataset-stats}
\end{table}

\section{Experimental results}
\label{sec:results}
In this section, we present the \ac{osgr} results obtained with \ac{pcaa} on our dataset of radar point clouds.

\subsection{Evaluation metrics}
\label{sec:evaluation-metrics}
The difficulty of open-set classification problems depends on the ratio between the number of unknown and known subjects, $|S_U|/|S_K|$. The larger this ratio, the higher the difficulty since the \ac{osgr} model faces a larger number of unknown subjects with respect to the ones used for training. 
In~\cite{geng2020recent} the authors propose the \textit{openness} value to express the \ac{osgr} difficulty with a quantity in the range $\left[0,1\right)$. Openness is defined as
\begin{equation}
    \label{eq:openness}
    \text{Openness} = 1 - \sqrt{\frac{2 |S_K|}{2|S_K| + |S_U|}},
\end{equation}
where values closer to $1$ represent higher difficulty. In the following, we express the openness in $\%$.

Evaluating the performance of \ac{osgr} models requires carefully choosing a suitable performance metric. 
Indeed, given a dataset, the learning problem is affected by class imbalance depending on the considered level of openness. This is true even for datasets that are balanced when considering an openness equal to $0$, i.e., when the dataset is used for a standard closed-set classification task. 
For instance, at low openness, the unknown partition $\mathcal{U}$ contains much fewer samples than the known partition $\mathcal{K}$. We observe instead an opposite imbalance when the openness is high.

To address this problem and provide a fair performance evaluation, we aggregate the F$1$~score of each class using the \textit{macro} aggregation strategy~\cite{manning2008introduction}. 
With this approach, the predictive accuracy for each class is weighted equally, regardless of the number of samples in each class. 
The macro aggregation strategy calculates the F$1$ score as follows
\begin{equation}
\text{F$1$} = \frac{1}{N} \sum_{i=1}^{N} \frac{2 \text{TP}_i}{2 \text{TP}_i + \text{FP}_i + \text{FN}_i}
\end{equation}
where $N = |S_{\mathcal{K}}|+1$ is the total number of classes, and $\text{TP}_i$, $\text{FP}_i$, and $\text{FN}_i$ represent the true positives, false positives, and false negatives for class $i$, respectively.


\subsection{Performance evaluation}
\label{sec:sota-comparison}

In this section, we compare \ac{pcaa} to existing solutions in terms of predictive performance. 

\noindent\textit{Methodology and choice of the baselines}:
To the best of our knowledge, our approach is the first to tackle the OSGR task from spatio-temporal point clouds. 
The lack of existing approaches for point-cloud-based OSGR poses a challenge in defining a fair and direct comparison with other established baselines. \revv{As a first step, we identify the most recent state-of-the-art approaches for \ac{osgr}, namely \ac{orced}~\cite{yang2022multiscenario} and \ac{lgm}~\cite{ni2022gait}, to use as baselines for our experiments. As discussed in \secref{sec:related-work-osgr}, these methods are designed for \ac{md} spectrograms. Therefore, to test these approaches on the same data modality for which they were designed, we also obtain \ac{md} spectrograms from our radar measurements and use them as training and testing data. However, to distinguish the impact of the different data representation (\ac{md} vs. point clouds) from that of the different architecture, we also test the performance of the baselines under the different data modality of point clouds, which is the main focus of this work. Indeed, the processing of point clouds entails additional challenges, such as temporal correlation modeling and ensuring the invariance of the output to input points permutations. Therefore, we decided to introduce a variation of each baseline approach, adapted for point-cloud processing. We will denote these two variants of \ac{orced} and \ac{lgm} as \mbox{PC-\ac{orced}} and \mbox{PC-\ac{lgm}}, respectively.}
We kept the modifications to the minimum needed to extend the baselines to point cloud data, to provide the fairest possible comparison. 
\revv{By testing the baselines both on spectrogram images and on spatio-temporal point clouds, we enable a comprehensive comparison that accounts both for architectural and data modality differences.}

Besides the minimal needed changes, these point-cloud variants preserve the high-level architectural components of \ac{orced} and \ac{lgm} without changes, such as loss functions, regularization strategies, and overall network structure. More in detail:

\begin{itemize}
    \item The original \ac{orced} implements a probabilistic ladder architecture~\cite{sun2020conditional}~\cite{sonderby2016train} with a symmetric encoder-decoder structure, which was designed for the processing of micro-Doppler spectrograms and is not directly applicable to point cloud sequences. 
    \revv{Therefore, in \mbox{PC-\ac{orced}}, we replace that part of the network with the Encoder and Decoder modules of \ac{pcaa}, which are suitable for point clouds processing.}
    \item On the other hand\revv{, for PC-\ac{lgm},} we adapt the original approach by first reproducing the same network structure as in \cite{ni2022gait}, but replacing the ResNet-18 feature extractor backbone, that is only suitable for image data, with the Encoder block of \ac{pcaa}, similarly to what we do for \revv{\mbox{PC-\ac{orced}}}. We then train the adapted network using the same original loss function. 
\end{itemize}
Notably, our only modifications to the baselines concern the feature extraction network that operates on point clouds.
Hence, our comparisons highlight high-level architectural features rather than the implementation of the individual modules, due to the different nature of the input data types involved (point cloud sequences and \ac{md} spectrograms).

More specifically:
\begin{enumerate}
    \item \ac{pcaa} leverages the same approach of \ac{aae}, thus employing a dedicated adversarial module (the discriminator) to enforce a regularized distribution of the latent space. 
    \rev{\ac{orced} and \ac{lgm} also aim at regularizing the latent space, but using different strategies, namely the \ac{vae} architecture and the \ac{lgm} loss function}, respectively.
    \item \ac{orced} adopts an additional \revv{contrastive loss} term applied to vectors in latent space, which we do not use. This is because the desired clustering effect around the class centroids is intrinsically obtained by \ac{pcaa} during the regularization phase.
    \item While \ac{orced} \rev{and \ac{lgm} learn} the parameters of the class distributions in the latent space, in \ac{pcaa} we fix the parameters and allow the network to focus on how the latent vectors should be arranged in the latent space, given the prior distribution.
\end{enumerate}
\revv{
}

\begin{figure}
    \centering
    \includegraphics[width=\linewidth, trim={0 0 0 1.5cm}, clip]{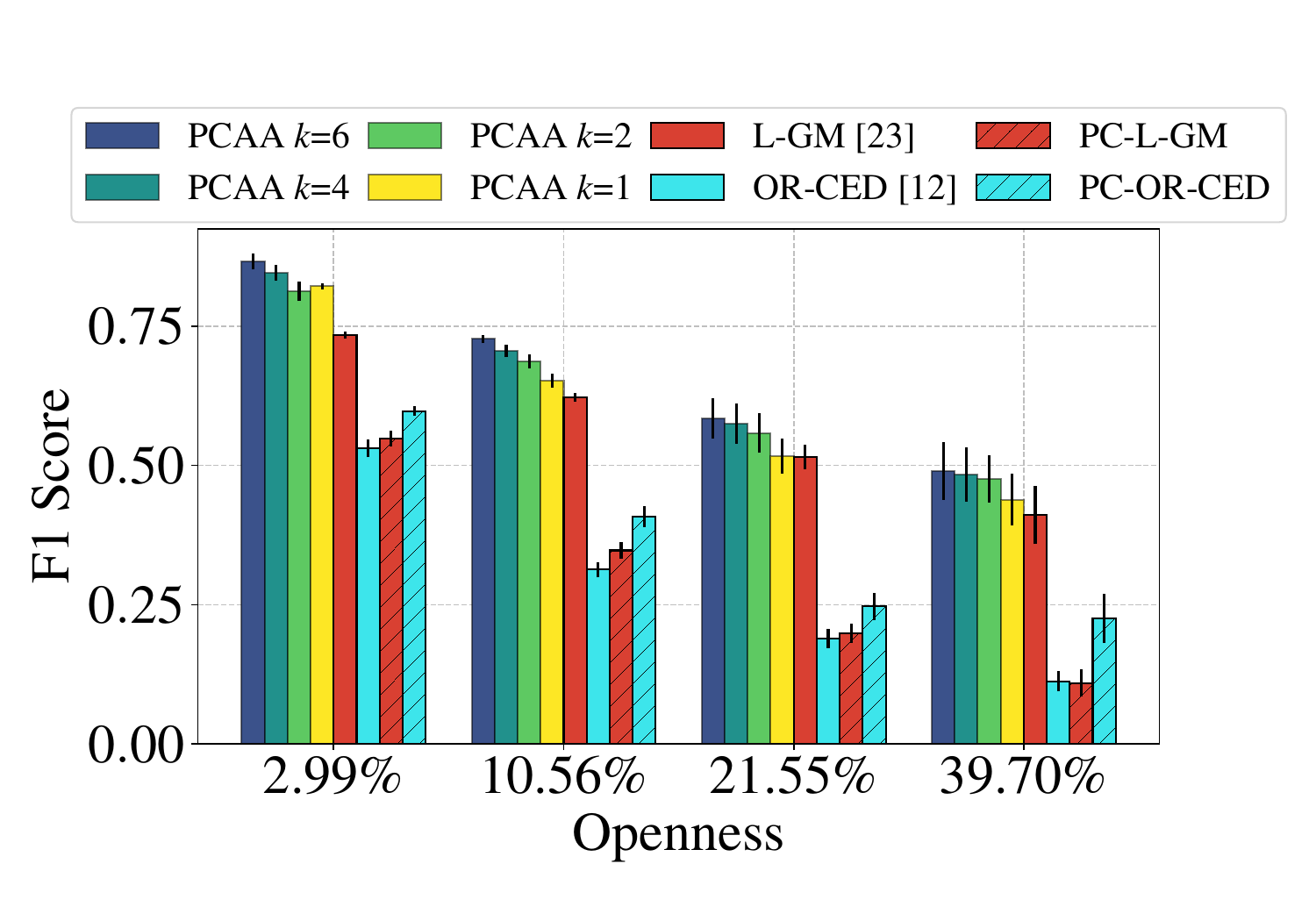}
    \caption{\revv{\ac{osgr} macro-averaged F1 scores of \ac{orced}, \ac{lgm}, \ac{pcaa}, \mbox{PC-\ac{orced}} and \mbox{PC-\ac{lgm}}, for different levels of openness. We also vary the number of aggregation steps $k$.}}
    \label{fig:sota_barplot}
\end{figure}

\noindent\textit{Results discussion}: We evaluate the performance of all models under different levels of openness to test their robustness to more challenging scenarios. We also report the performance of \ac{pcaa} with different values of $k$ for the novel subject detection phase. 
\fig{fig:sota_barplot} shows the performance of \ac{pcaa} against \rev{the benchmarks.}
We average our results over $n_{\text{test}}=5$ different trials, each employing a different dataset partitioning $(\mathcal{K}, \mathcal{U})$. 
We report the mean F$1$-score value along with a measure of the dispersion of the results, obtained as $\sigma/\sqrt{n_{\text{test}}}$, where $\sigma$ is the standard deviation of the F$1$-scores over the $n_{\rm test}$ trials.
We highlight that the randomization of the partitions is key to ensuring a fair evaluation. 
Indeed, the difficulty of the unknown detection is heavily influenced by \textit{which} subjects are present in the unknown set, due to the individual variability of the gait features.

\rev{Additionally, our approach exhibits a clear trend of improved performance as the value of $k$ increases. This behavior represents one of the key advantages of our method over existing ones, as it enables users to balance inference speed and accuracy in the \ac{osgr} task by adjusting $k$ accordingly. For instance, in scenarios where only a few unknowns are expected, the openness of the problem can be assumed to be low, allowing small values of $k$ to yield both fast and accurate results. Conversely, in more challenging settings, where the number of unknowns is large, higher values of $k$ can be selected to reduce misclassifications at the cost of increased computational time.
}

\begin{figure}[t]
    \centering
    \begin{subfigure}[b]{0.24\textwidth}
        \centering
        \includegraphics[width=\textwidth]{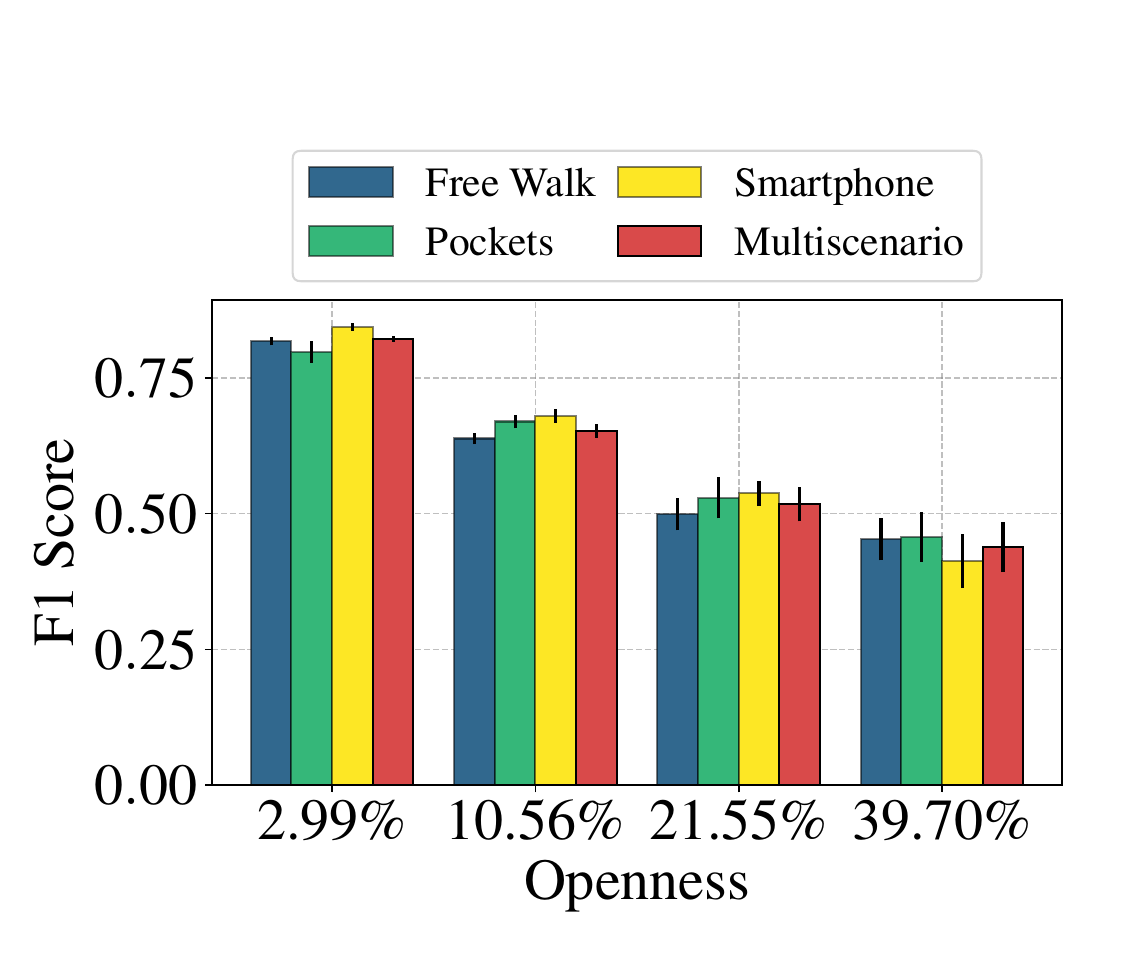}
        \caption{Results for $k=1$}
        \label{fig:scenarios_plot01}
    \end{subfigure}%
    \hfill
    \begin{subfigure}[b]{0.24\textwidth}
        \centering
        \includegraphics[width=\textwidth]{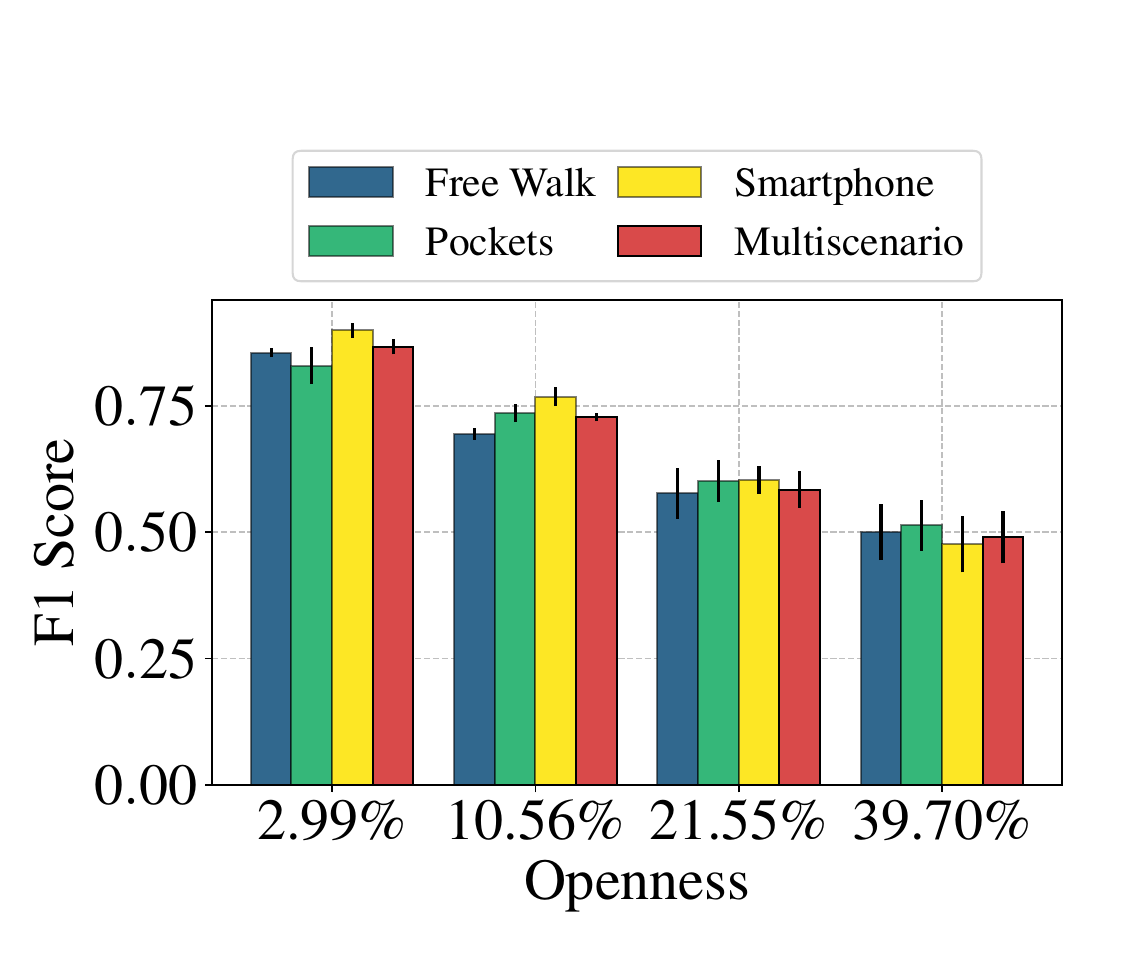}
        \caption{Results for $k=6$}
        \label{fig:scenarios_plot02}
    \end{subfigure}%
 
    \caption{Performance for separate scenarios.}
    \label{fig:plot_scenarios}
\end{figure}

\begin{figure}[t]
    \centering
    \begin{subfigure}[b]{0.24\textwidth}
        \centering
        \includegraphics[width=\textwidth]{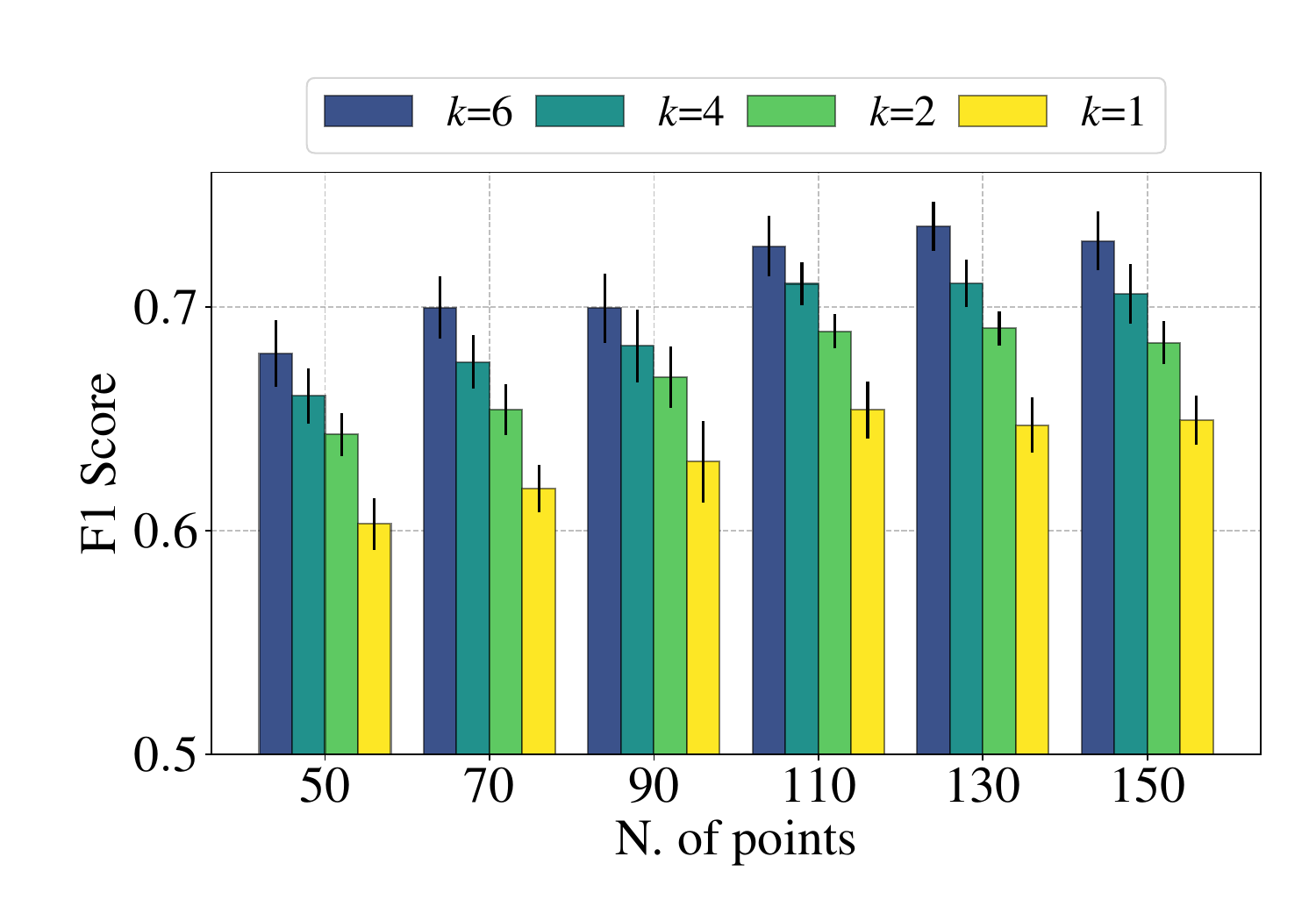}
        \caption{Openness $10.56\%$}
        \label{fig:npoints01}
    \end{subfigure}%
    \hfill
    \begin{subfigure}[b]{0.24\textwidth}
        \centering
        \includegraphics[width=\textwidth]{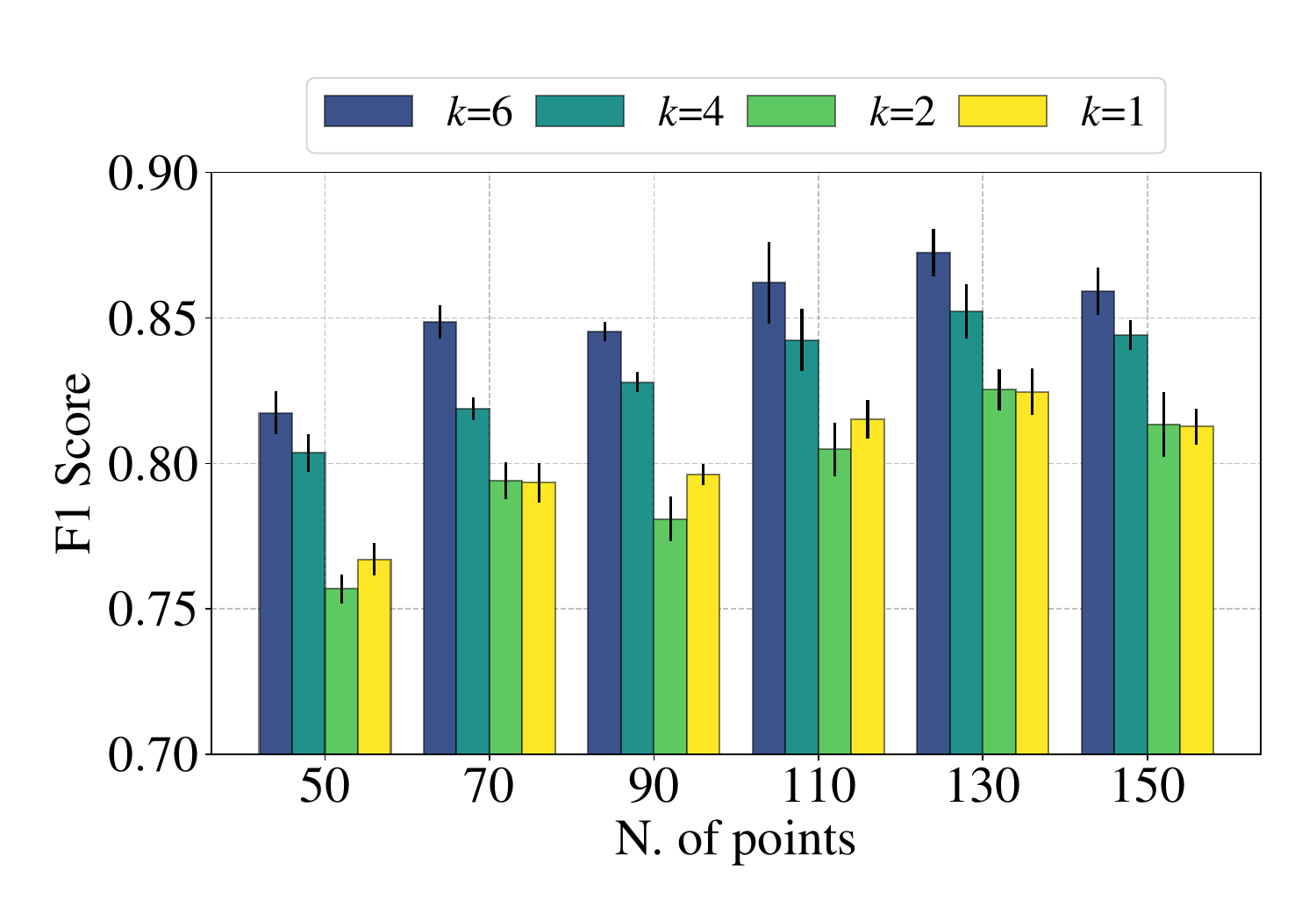}
        \caption{Openness $2.99\%$}
        \label{fig:npoints02}
    \end{subfigure}%
 
    \caption{Performance for different number of points in the point clouds.}
    \label{fig:plot_density}
\end{figure}

In \fig{fig:plot_scenarios} we also report the performance of \ac{pcaa}, trained with the whole dataset, for each isolated walking modality. 
This is done to provide an estimate of which might be the most challenging and most informative walking modality during the inference phase. 
The results suggest that subjects walking while using a smartphone are slightly more easily distinguished with respect to the other two walking modalities, across almost all the levels of problem openness.
As part of our experiments, we also explored the role of the maximum number of points in the input point clouds. 
In \fig{fig:plot_density} we report the performance of \ac{pcaa}, when trained and tested on a varying number of input points. 
As expected, higher values of point cloud density lead to increased performance on the final \ac{osgr} task. 
It is interesting to observe, however, that the performance variation is not significant for point clouds with more than $130$ points, and it also leads to a slight decrease in performance in the case of smaller problem openness. 

It is important to take into account that \rev{both benchmarks perform}
novel subject detection using a single input sample. Due to the high variability of human gait patterns, a single sample is sometimes not enough to differentiate among subjects. Conversely, our method is capable of waiting for $k\geq1$ input samples before returning a prediction, observing a longer point cloud sequence to improve its confidence. To provide a fair comparison, we also report results for \ac{pcaa} for $k=1$.
It is clear that the complexity of the problem significantly increases with the openness for all the considered approaches.
This causes a general decreasing trend in the F$1$ score.

\begin{figure*}[t]
    \centering
    \begin{subfigure}[b]{0.22\textwidth}
        \centering
        \includegraphics[width=\textwidth]{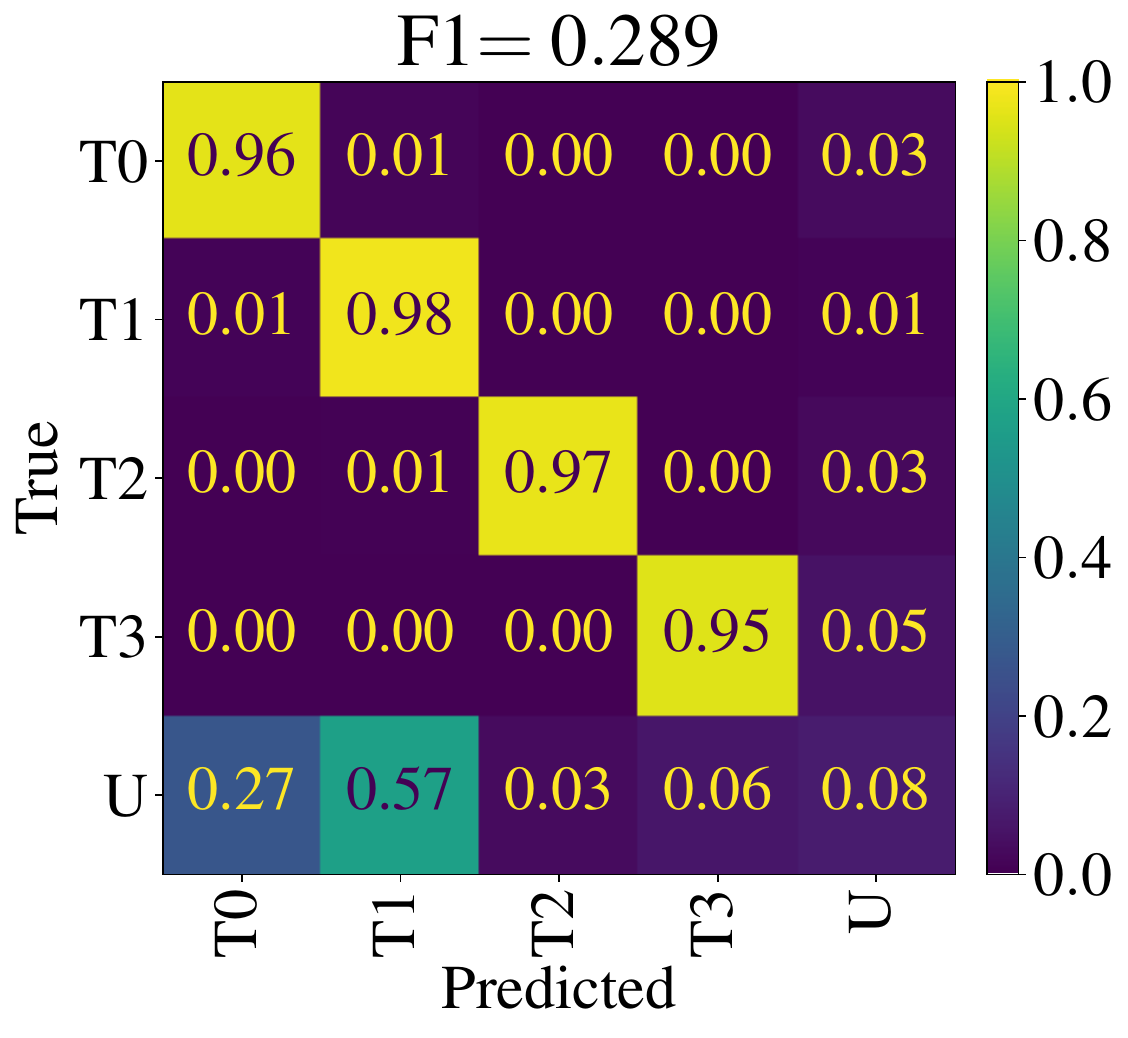}
        \caption{PC-\ac{orced}}
        \label{fig:image1}
    \end{subfigure}%
    \hfill
    \begin{subfigure}[b]{0.22\textwidth}
        \centering
        \includegraphics[width=\textwidth]{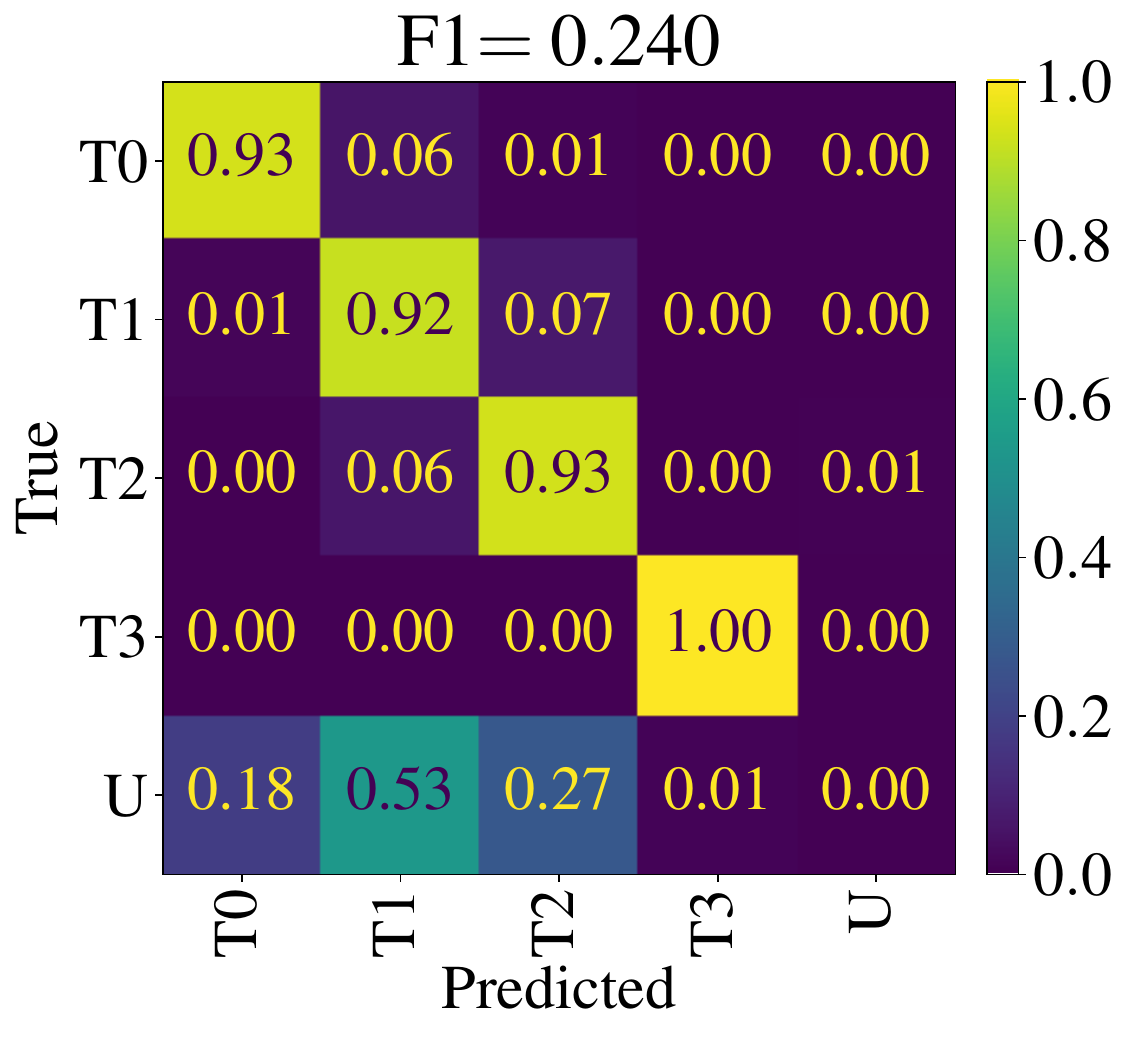}
        \caption{\rev{PC-\ac{lgm}}}
        \label{fig:image1}
    \end{subfigure}%
    \hfill
    \begin{subfigure}[b]{0.22\textwidth}
        \centering
        \includegraphics[width=\textwidth]{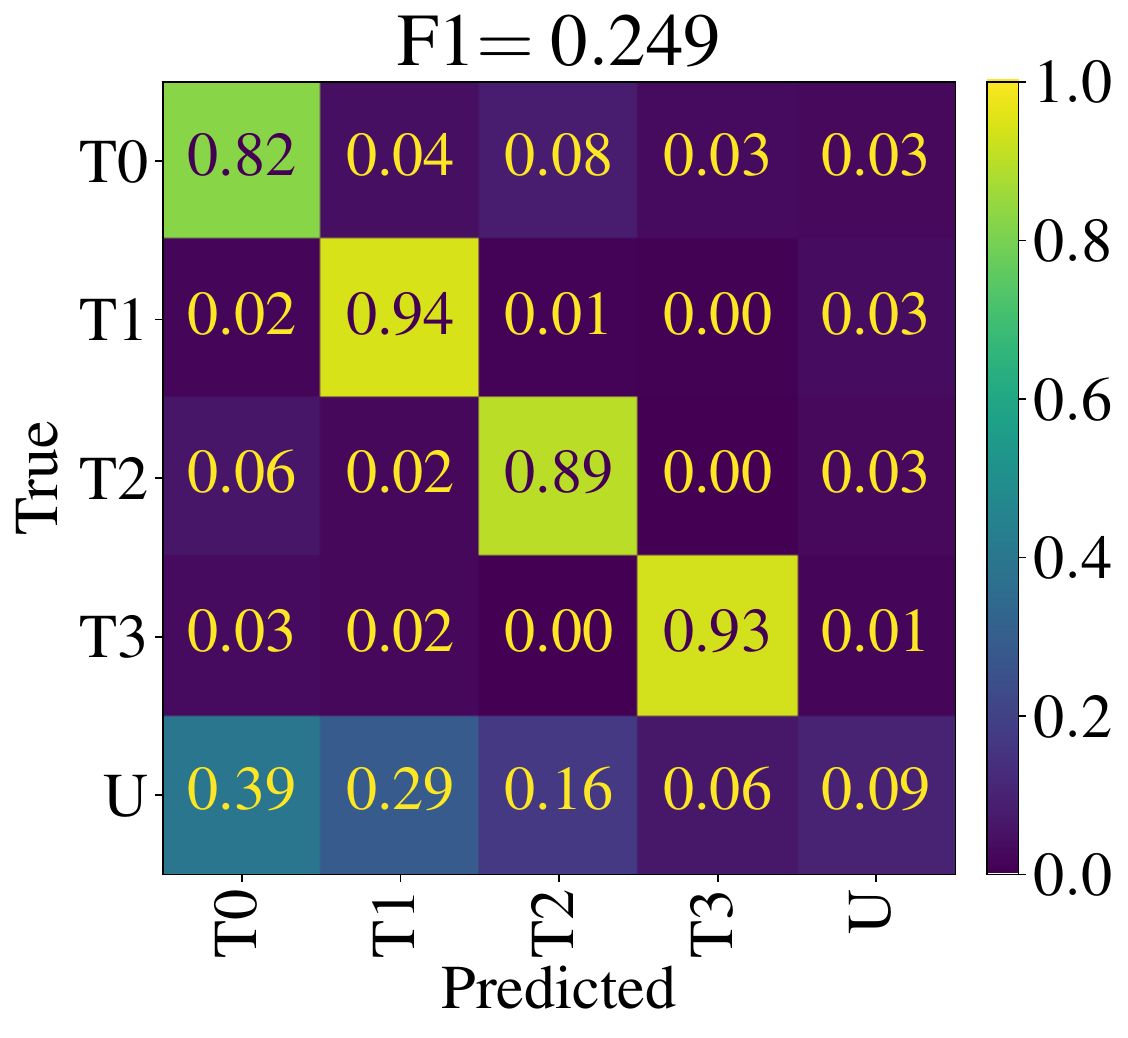}
        \caption{\ac{orced} \cite{yang2022multiscenario}}
        \label{fig:image2}
    \end{subfigure}%
    \hfill
    \begin{subfigure}[b]{0.22\textwidth}
        \centering
        \includegraphics[width=\textwidth]{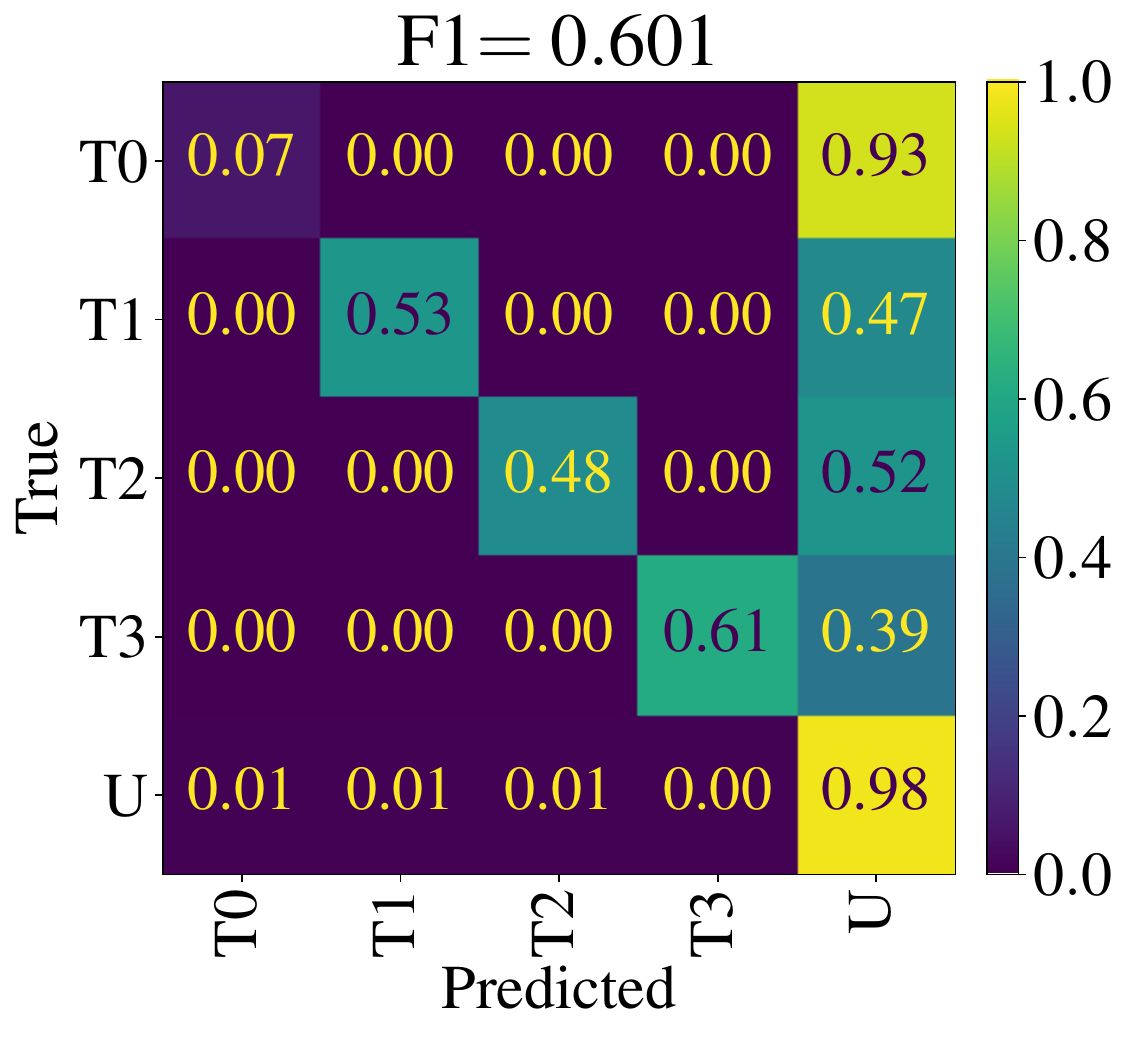}
        \caption{\ac{lgm} \cite{ni2022gait}}
        \label{fig:image3}
    \end{subfigure}%
    \hfill
    \\[\baselineskip]
    \hfill
    \begin{subfigure}[b]{0.22\textwidth}
        \centering
        \includegraphics[width=\textwidth]{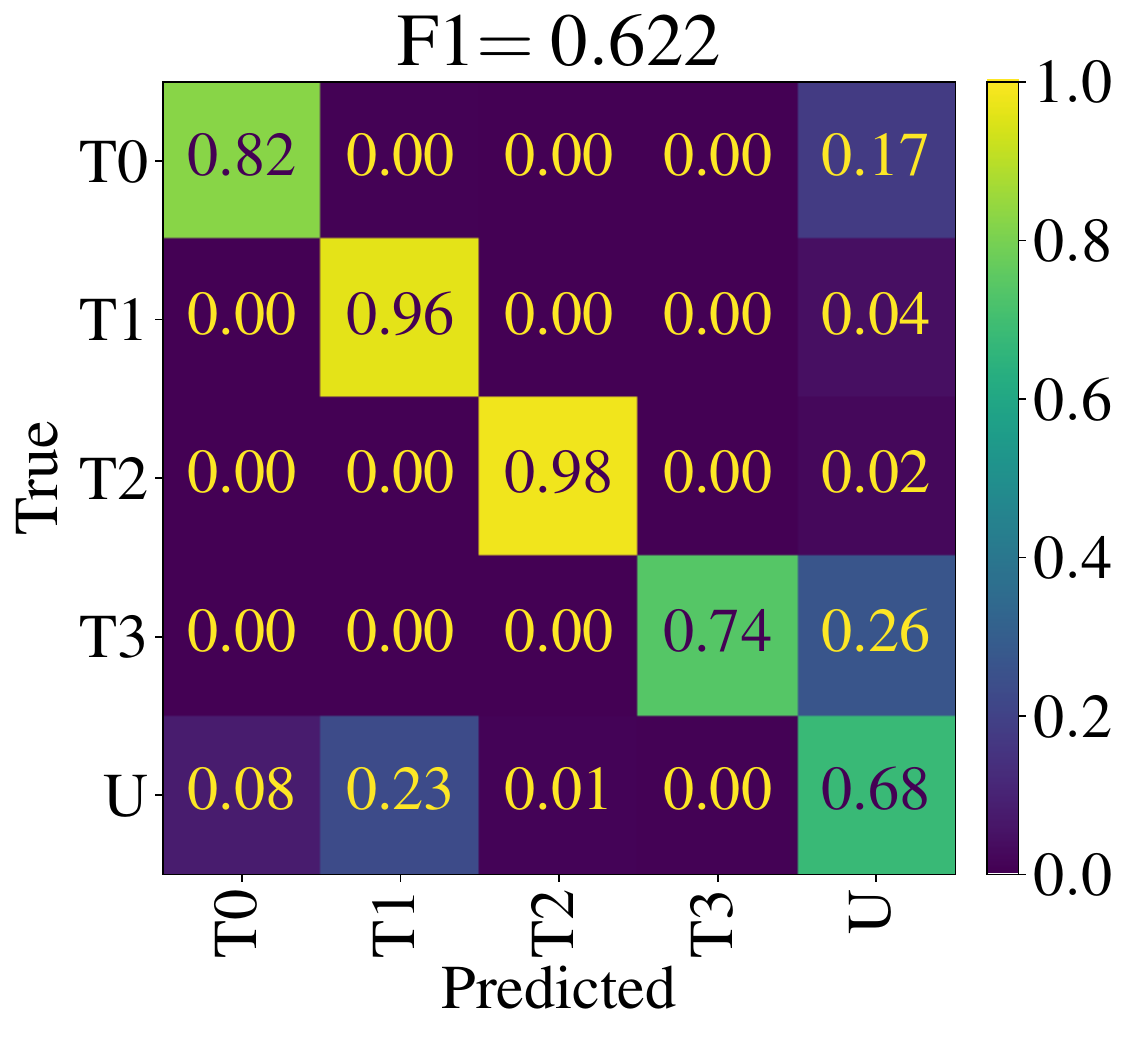}
        \caption{\ac{pcaa} with $k=1$}
        \label{fig:image2}
    \end{subfigure}%
    \hfill
    \begin{subfigure}[b]{0.22\textwidth}
        \centering
        \includegraphics[width=\textwidth]{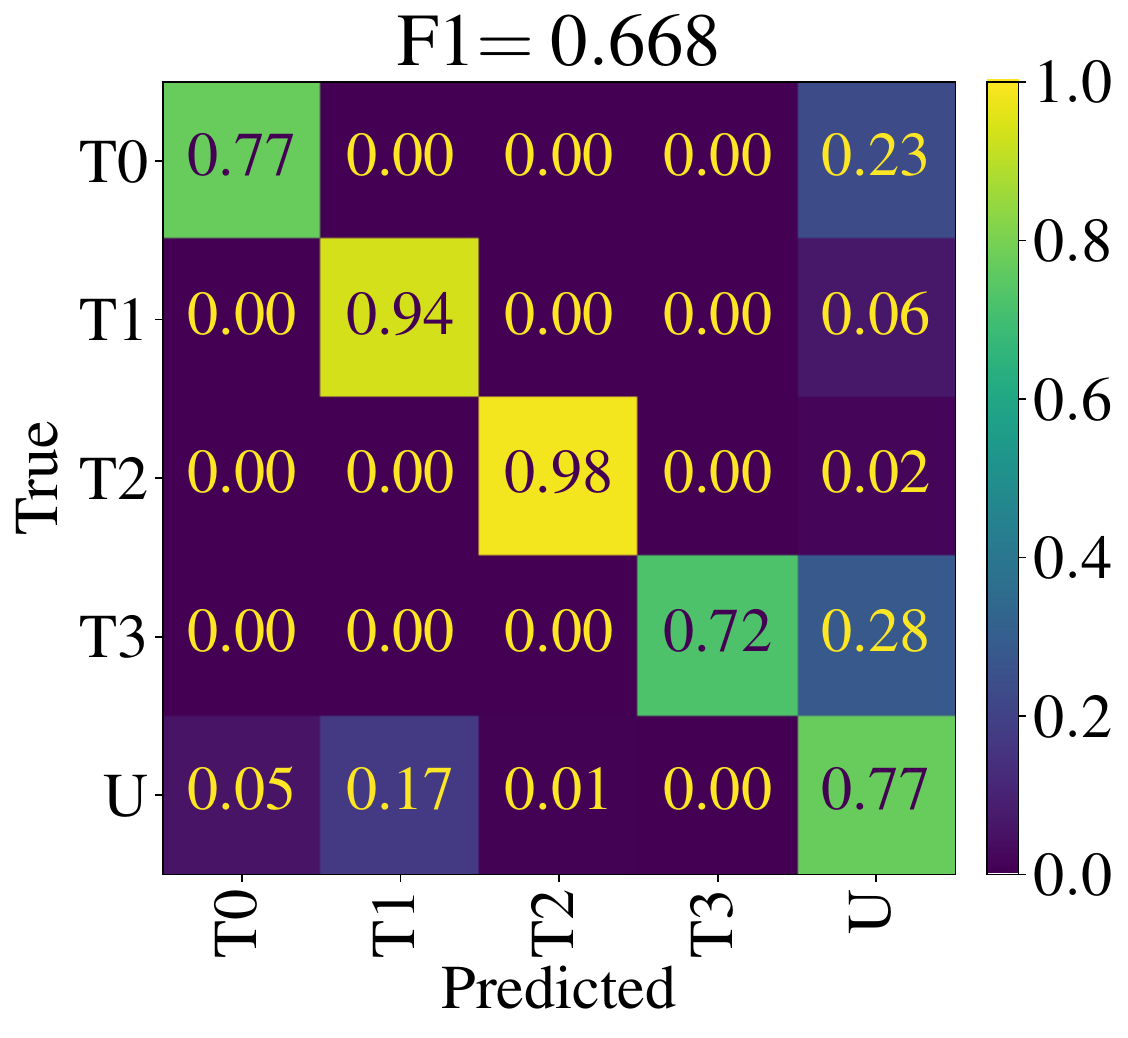}
        \caption{\ac{pcaa} with $k=2$}
        \label{fig:image3}
    \end{subfigure}%
    \hfill
    \begin{subfigure}[b]{0.22\textwidth}
        \centering
        \includegraphics[width=\textwidth]{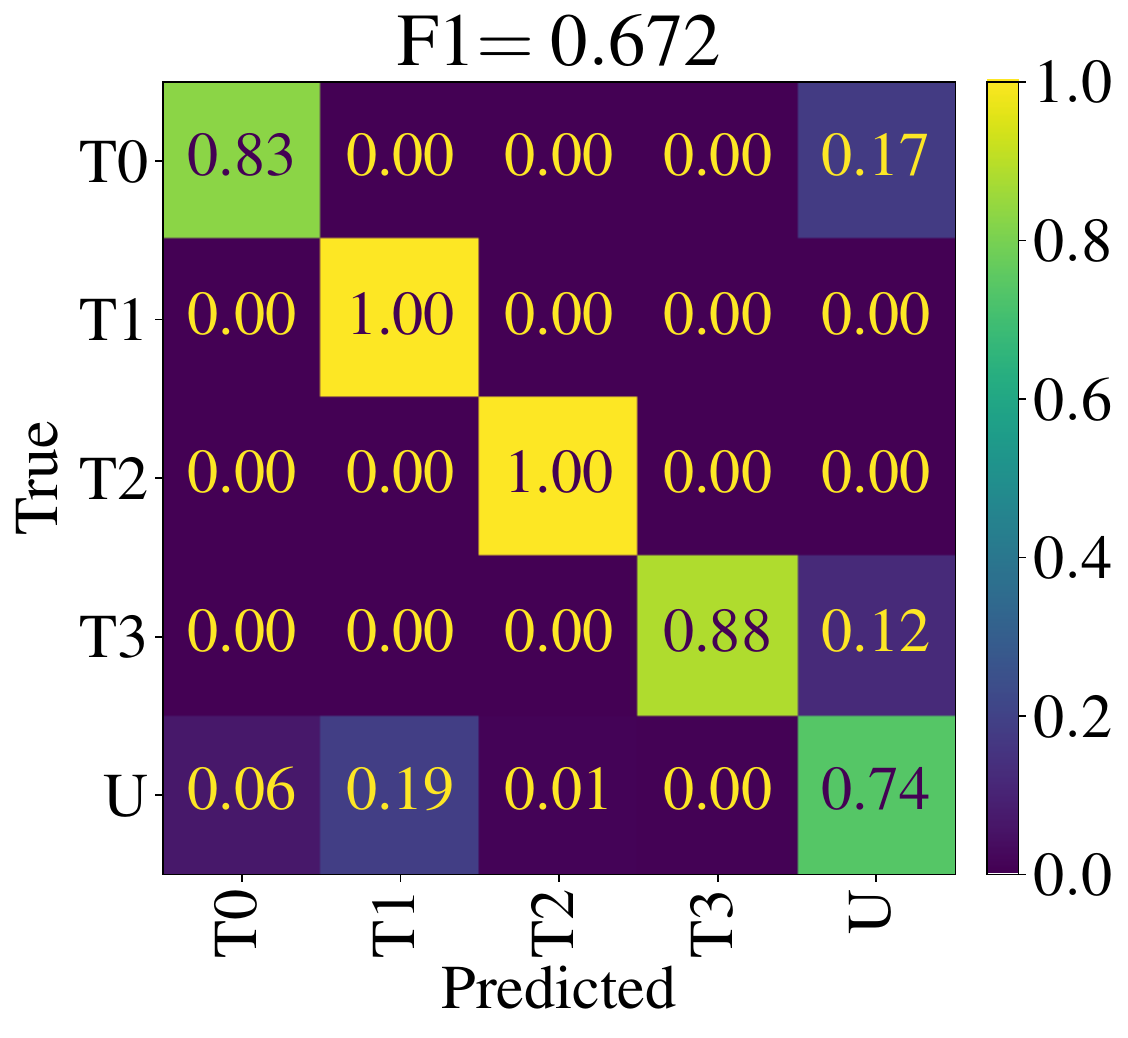}
        \caption{\ac{pcaa} with $k=4$}
        \label{fig:image4}
    \end{subfigure}
    \hfill
    \begin{subfigure}[b]{0.22\textwidth}
        \centering
        \includegraphics[width=\textwidth]{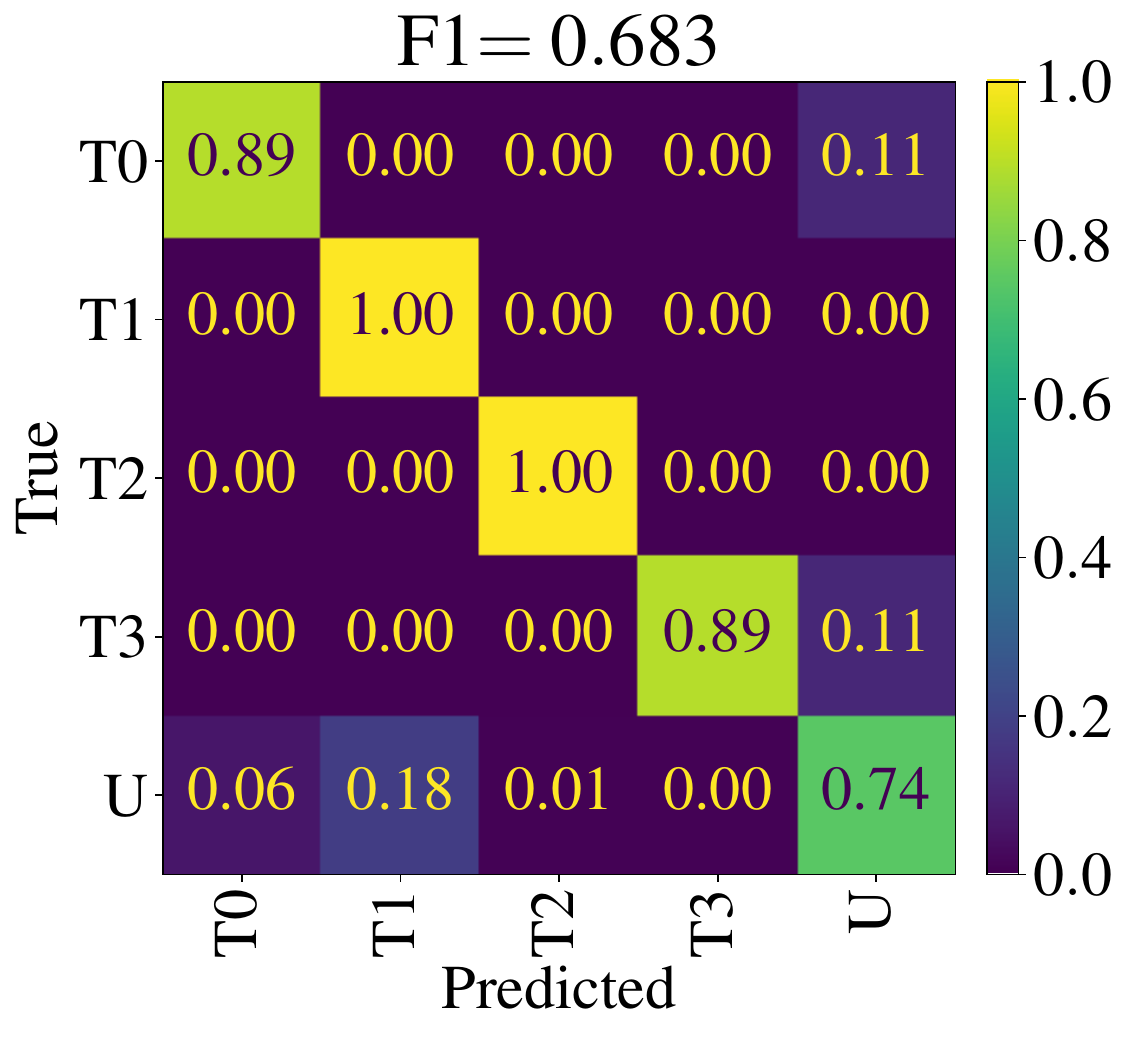}
        \caption{\ac{pcaa} with $k=6$}
        \label{fig:image1}
    \end{subfigure}%

    \caption{Example of confusion matrices in a scenario with $21.55\%$ openness (9 subjects of which 4 known and 5 unknown). 
    \revv{\ac{orced}, \mbox{PC-\ac{orced}} and \mbox{PC-\ac{lgm}} perform very well on the closed-set recognition task, but are not able to detect novel samples. When trained and tested on spectrogram images, \ac{lgm} performs significantly better, but still often mistakes known subjects for unknowns.} Conversely, \ac{pcaa} can distinguish the majority of unknown subjects\revv{, and its performance significantly improves as $k$ increases.}}
    \label{fig:conf_matrices}
\end{figure*}

\revv{\fig{fig:sota_barplot} also shows the performance of the baselines, \ac{orced} and \ac{lgm}, trained and tested both with \ac{md} spectrogram data, as well as \mbox{PC-\ac{orced}} and \mbox{PC-\ac{lgm}}, trained and tested on our sequential point cloud data. There is a clear performance drop between \ac{lgm} and \mbox{PC-\ac{lgm}}, which shows that models designed for \ac{md} spectrograms are not suited for point cloud data. \ac{orced} performs poorly in both data representations. This could be attributed to the differences in data distribution between our dataset and the dataset employed in \cite{yang2022multiscenario}. Notably, \ac{pcaa} outperforms all the considered baselines, even when trained and tested employing their native data representations. To better showcase the performance of each considered approach, we report some examples of confusion matrices in \fig{fig:conf_matrices}}


\revv{Our results demonstrate the superiority of \ac{pcaa} over existing approaches on the task of \ac{osgr} from spatio-temporal point clouds. Our results also confirm that adapting existing methods for \ac{osgr} to support point clouds processing is \textit{insufficient} to achieve the same level of performance. This further underscores the complexity of solving the \ac{osgr} task from radar point cloud sequences.}

\rev{
Additionally, to demonstrate the computational and memory efficiency of our approach, we report the observed metrics regarding training and inference times, and memory usage in \tab{tab:efficiency_vs_k}.
}

\begin{table}[ht]
\centering
\caption{\rev{Time and memory requirements for \ac{pcaa}. Training times are reported for different values of $S_K$, the number of subjects present in the training set. Inference times and input sizes are reported assuming input batches of $k$ samples, for different values of $k$.
}}
\begin{threeparttable}
{\begin{tabular}{lcr}
\toprule
\textbf{Metric [unit]} & \textbf{Case} & \textbf{Value} \\
\midrule
\multirow{4}{*}{Training time [min]}             & $S_K=2$ &  $28.83 \pm 0.66$ \\
                                         & $S_K=4$ & $57.15 \pm 0.68$ \\
                                         & $S_K=6$ & $86.41 \pm 2.02$ \\
                                         & $S_K=8$ & $119.58 \pm 3.06$ \\


\midrule
\multirow{4}{*}{Inference time [ms]}            & $k=1$ & $2.76 \pm 0.43$ \\
                                           & $k=2$ & $5.10 \pm 0.73$ \\
                                           & $k=4$ & $9.63 \pm 1.31$ \\
                                           & $k=6$ & $14.02 \pm 1.82$ \\
\midrule
\multirow{4}{*}{Input size [MB]}    & $k=1$ & $0.069$ \\
                                           & $k=2$ & $0.137$ \\
                                           & $k=4$ & $0.275$ \\
                                           & $k=6$ & $0.412$  \\
\midrule
\multirow{3}{*}{Model size [MB]}   & Encoder &  $9.305$ \\
                            & Decoder\tnote{*}  &  $821.155$ \\
                            & Discriminator\tnote{*}  &  $0.017$ \\

\bottomrule

\end{tabular}}
\begin{tablenotes}\footnotesize
\item[*] \rev{Not used during inference}
\end{tablenotes}
\end{threeparttable}
\label{tab:efficiency_vs_k}
\end{table}

\rev{
The reported values show that our approach exhibits negligible inference times, i.e., in the order of milliseconds, if compared to the actual duration of the input point cloud sequences, which include a few seconds of human gait information. This holds even for higher values of $k$. Moreover, \ac{pcaa} has a remarkably low memory footprint, both in terms of model parameters and input point cloud representations. These features make our approach suitable for edge computing applications involving resource-constrained devices.
}

\subsection{Ablation studies}
\label{sec:ablation}

To further validate the impact of each component of \ac{pcaa}, we conducted a series of ablation studies. To this end, we designed and implemented three versions of the proposed method. 

\textbf{V1:} In this version, the class centroids $\boldsymbol{\mu}_i$ defined in \secref{subsec:model-architecture} are learnt by the network instead of being fixed. More specifically, we replace the deterministic mapping $f$ with a multi-layer perceptron with $3$ hidden layers of dimension $(16, 32, 64)$. 

\textbf{V2:} In this version, we remove the projection heads $\mathcal{H}_D$ and $\mathcal{H}_C$. This is done to assess if these projection heads are responsible for providing more specialized features for each branch of the network, ensuring that the features coming from latent space remain as general as possible.

\textbf{V3:} In the third version, we remove the Decoder block from the architecture. This is done to demonstrate the necessity of the unsupervised learning branch in the regularization of the latent space.

The results of our ablation studies are reported in \tab{tab:ablation_studies}. We report the mean and standard deviation of the results across five different runs.

\begin{table*}[h]
\small
\centering
\caption{Ablation studies results in terms of F$1$-Score.}
\label{tab:ablation_studies}
\begin{tabular}{ccccc}
\toprule
\multirow{2}{*}{\textbf{Method}} & \multicolumn{4}{c}{\textbf{Openness}} \\
 & 2.99\% & 10.56\% & 21.55\% & 39.7\% \\ 
 \cmidrule(lr){1-1} \cmidrule(lr){2-5}
V1 & $0.239 \pm 0.214$ & $0.198 \pm 0.095$ & $0.223 \pm 0.061$ & $0.313 \pm 0.146$ \\
V2 & $0.871 \pm 0.033$ & $\mathbf{0.762 \pm 0.025}$ & $0.562 \pm 0.192$ & $0.392 \pm 0.100$ \\
V3 & $0.727 \pm 0.057$ & $0.507 \pm 0.023$ & $0.331 \pm 0.098$ & $0.300 \pm 0.088$\\
\textbf{\ac{pcaa}} & $\mathbf{0.874\pm0.025}$ & $0.730 \pm 0.016$ & $\mathbf{0.587 \pm 0.088}$ & $\mathbf{0.492 \pm 0.118}$ \\
\bottomrule

\end{tabular}
\end{table*}

Across three out of four levels of openness, \ac{pcaa} outperforms all other variants, proving the importance of each architectural choice. V2 achieves the second-highest F$1$ score, slightly surpassing \ac{pcaa} at one level of openness. 

In contrast, V1 performs significantly worse than the other variants, clearly indicating that fixing the class centroids represents a critical design choice. Similarly, V3 demonstrates significantly worse performance compared to both V2 and \ac{pcaa}, highlighting the critical role of the Decoder in solving the \ac{osgr} task. In contrast, the removal of the projection head modules has only a marginal effect on the results, except in the scenario with the highest Openness.

\section{Concluding Remarks}
\label{sec:conclusion}

In this work, we address open-set gait recognition from sparse radar point cloud sequences.
We introduce \ac{pcaa}, an original dual-branch neural network architecture designed to simultaneously classify subjects seen during training and generate a well-regularized, semantically rich feature space of human gait patterns. Finally, we solve the \ac{osgr} task by applying a novelty detection algorithm that exploits the structure and regularity of the latent space. 
Unlike other approaches in the literature, our novel subject detection algorithm accumulates evidence across time before providing an answer, resulting in increased detection accuracy while retaining acceptable inference time. 
This feature makes our approach flexible, allowing the user to trade accuracy for faster prediction times or vice versa, depending on the openness of the problem.

To address the lack of publicly available high-quality data for this task, we collect and publicly release \datasetname{}, a radar dataset containing \revv{both point cloud traces and \ac{md} spectrograms of} $10$ human subjects, each \revv{walking} with three distinct walking modalities.
Using \datasetname{}, we compare our approach to the 
\rev{most recent}
existing state-of-the-art \rev{approaches for \ac{osgr}, \ac{orced} and \ac{lgm}.}
\revv{To provide a comprehensive and fair comparison, we evaluated the performance of the baselines both on \ac{md} spectrograms (the data type they were designed for) and on point cloud data, which is the focus of this work.}

Our results show the superiority of our approach in terms of \ac{osgr} F$1$ score in different settings and openness levels, \rev{clearly confirming that a simple adaptation of existing gait recognition strategies for \ac{md} signatures is not sufficient to achieve the same performance when using sparse and noisy point clouds}. 
We additionally validate each architectural choice, by performing an extensive ablation study involving three variants of \ac{pcaa}.
Our ablation studies clearly show that the architectural features of \ac{pcaa}, tailored for point cloud input sequences, are crucial in our setting.
This highlights the necessity of specifically designing architecture blocks and learning strategies to deal with complex data structures like \rev{spatio-temporal} radar point clouds.

\bibliographystyle{IEEEtran}
\bibliography{IEEEabrv,biblio}

\vfill

\newlength{\biocompact}
\setlength{\biocompact}{-10mm}

\end{document}